\documentclass[twocolumn]{svjour3}          % twocolumn
\pdfoutput=1
\usepackage{graphics }
\usepackage{longtable}
\usepackage{times}
\usepackage{verbatim}
\usepackage {url}
\usepackage{float}
\restylefloat{table}
\usepackage{float}
\floatstyle{plain}\restylefloat{figure}
\floatstyle{plaintop}\restylefloat{table}
\usepackage{placeins}
\usepackage{amssymb}
\usepackage{bm}
\usepackage{relsize}
\usepackage{esvect}
\usepackage{moreverb}
\usepackage[]{amsmath}
\usepackage[boxed,commentsnumbered]{algorithm2e}
\usepackage{array}
\usepackage{booktabs}
\usepackage{multirow}
\usepackage{hhline}
\usepackage[numbers]{natbib}
\usepackage{fancyhdr}
\usepackage{amsmath}
\usepackage{mathtools}

\newcommand{\argmin}{\operatornamewithlimits{argmin}}

\usepackage{color}
\newcommand\newnotecommand[3]{%
\newcommand#1[1]{{\color{#3}\footnote{{\color{#3}#2:} ##1}}}}
\newnotecommand\tuomas{Tuomas}{blue}
\newnotecommand\sahar{Sahar}{green}
\newnotecommand\heikki{Heikki}{red}
\newcommand{\ve}[1]{\mathrm{\mathbf{#1}}}
\begin{document}

\title{Resolving Overlapping Convex Objects in Silhouette Images by Concavity Analysis and Gaussian Process%\thanks{Grants or other notes
}


\author{Sahar Zafari $^{*1}$        \and
    Mariia Murashkina$^{1}$  \and
    Tuomas Eerola$^{1}$ \and
    Jouni Sampo$^{2}$ \and
    Heikki K{\"a}lvi{\"a}inen$^{1}$ \and
    Heikki Haario$^{2}$ 
}


\institute{ [1]\at
 Machine Vision and Pattern Recognition Laboratory (MVPR),  
 School of Engineering Science,
Lappeenranta-Lahti University of Technology LUT, 
 P.O. Box 20, 53851 Lappeenranta \\
          \email{firstname.lastname@lut.fi}           %  \\
       \and
        [2]\at
Mathematics Laboratory, 
School of Engineering Science,
Lappeenranta-Lahti University of Technology LUT, 
P.O. Box 20, 53851 Lappeenranta
\email{firstname.lastname@lut.fi} 
}

\date{Received: date / Accepted: date}

\maketitle

\begin{abstract}
Segmentation of overlapping convex objects has various applications, for example, in nanoparticles and cell imaging. Often the segmentation method has to rely purely on edges between the background and foreground making the analyzed images essentially silhouette images. Therefore, to segment the objects, the method needs to be able to resolve the overlaps between multiple objects by utilizing prior information about the shape of the objects.
This paper introduces a novel method for segmentation of clustered partially overlapping convex objects in silhouette images. The proposed method involves three main steps: pre-processing, contour evidence extraction, and contour estimation. 
Contour evidence extraction starts by recovering contour segments from a binarized image by detecting concave points. After this, the contour segments which belong to the same objects are grouped. The grouping is formulated as a combinatorial optimization problem and solved using the branch and bound algorithm. Finally,  the full contours of the objects are estimated by a Gaussian process regression method. The experiments on a challenging dataset consisting of nanoparticles demonstrate that the proposed method outperforms three current state-of-art approaches in overlapping convex objects segmentation. The method relies only on edge information and can be applied to any segmentation problems where the objects are partially overlapping and have a convex shape.
\end{abstract}

\section{Introduction}\label{sec:1}
\begin{figure} [ht!]
\begin{center}
	\begin{tabular}{c c c}
		\frame{\includegraphics [width=.14\textwidth]{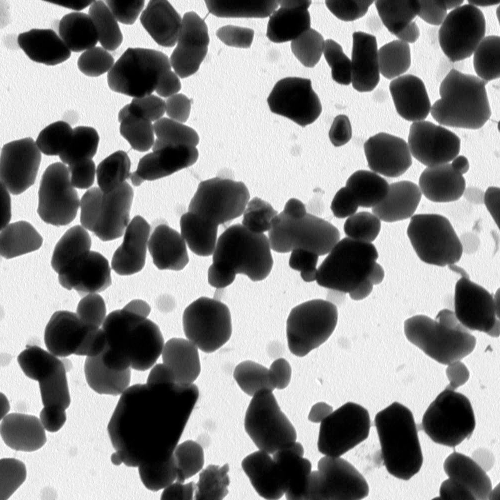}}&
		\frame{\includegraphics [width=.14\textwidth]{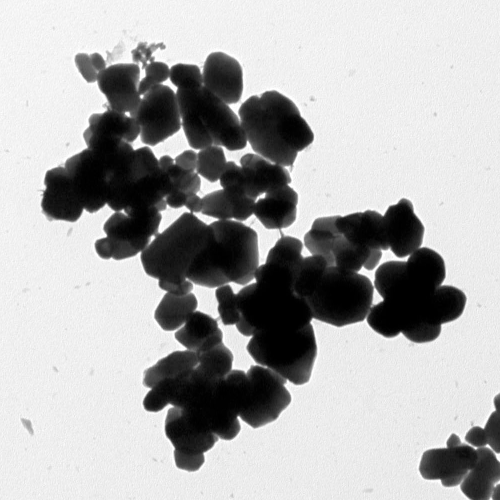}}&
		\frame{\includegraphics [width=.14\textwidth]{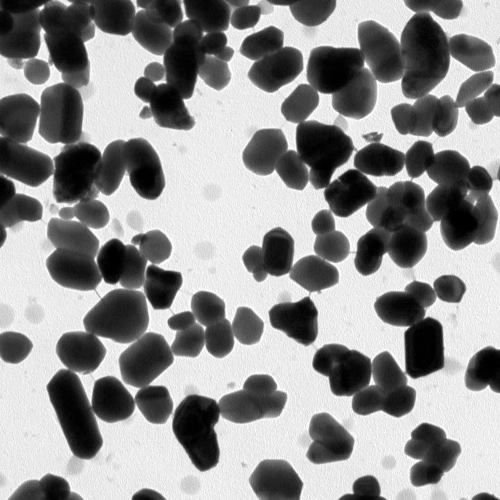}} 
	\end{tabular}
	%\vspace{.1cm}
	\caption{ Overlapping nanoparticles.} \label{fig:1}
\end{center}
\end{figure}

\sloppy Segmentation of overlapping objects aims to address the issue of representation of multiple objects with partial views. Overlapping or occluded objects occur in various applications, such as morphology analysis of molecular or cellular objects in biomedical and industrial imagery where quantitative analysis of individual objects by their size and shape is desired \cite{7300433, npa, bubble,5193169}. 
In these applications for further studies and analysis of the objects in the images, it is fundamental to represent the full contours of individual objects.
Deficient information from the objects with occluded or overlapping parts introduces considerable complexity into the segmentation process. For example, in the context of contour estimation, the contours of objects intersecting with each other do not usually contain enough visible geometrical evidence, which can make contour estimation problematic and challenging. Frequently, the segmentation method has to rely purely on edges between the background and foreground, which makes the processed image essentially a silhouette image (see Fig. \ref{fig:1}).
Furthermore, the task involves the simultaneous segmentation of multiple objects. A large number of objects in the image causes a large number of variations in pose, size, and shape of the objects, and leads to a more complex segmentation problem.
%----------------------------------% Figure 1 (example images)
%---------------------%

In this paper, a novel method for the segmentation of partially overlapping convex objects is proposed. The proposed method follows three sequential steps namely, pre-processing, contour evidence extraction, and contour estimation. 
The contour evidence extraction step consists of two sub-steps of contour
segmentation and segment grouping. In the contour segmentation, contours are divided into separate contour segments. In the segment grouping, contour evidences are built by joining the contour segments that belong to the same object. Once the contour evidence is obtained, contour estimation is performed by fitting a contour model to the evidences.

We, first, revise the comprehensive comparison of concave points detection
methods for the contour segmentation and the branch and bound (BB) based
contour segment grouping method from our earlier articles~\cite{Zafari2017} 
and~\cite{Zafari2017BB}, respectively. Then, we propose a novel 
Gaussian Process regression (GP) model for contour estimation. The proposed GP contour estimation method succeeded in obtaining the
more accurate contour of the objects in comparison with the earlier approaches.
Finally, we utilize these sub-methods to construct a full segmentation method for overlapping convex objects.

\sloppy The work makes two main contributions to the study of object segmentation. The first contribution of this work is the proposed GP model for contour estimation. 
The GP model predicts the missing parts of the evidence and estimates the full contours of the objects accurately. 
The proposed GP model for contour estimation is not limited to specific object shapes, similar to ellipse fitting, and estimates the missing contours based on the assumption that both observed contour points and unobserved contour points can be represented as a sample from a multivariate Gaussian distribution.
The second contribution is the
full method for the segmentation of partially overlapping convex objects. The proposed method integrates the proposed GP based contour estimation method with the previously proposed contour segmentation and segment grouping methods,
enabling improvements compared to existing segmentation methods with higher detection rate and segmentation accuracy.
The paper is organized as follows: Related work is reviewed in Sec.~\ref{sec:related}. Sec.~\ref{sec:method} introduces our method for the segmentation of overlapping convex objects. The proposed method is applied to a dataset of nanoparticle images and compared with four state-of-the-art methods in Sec.~\ref{sec:exp}. Conclusions are drawn in Sec.~\ref{sec:conclusion}.

\section{Related work\label{sec:related}}
\noindent
Several approaches have been proposed to address the segmentation of overlapping objects in different kinds of images~\cite{npa,ihc,sbfNuclei,bubble,5193169}. 
\sloppy The watershed transform is one of the commonly used approaches in overlapping cell segmentation \cite{ihc,4671118,5518402}. Exploiting specific strategies for the initialization, such as morphological filtering \cite{ihc} or the adaptive H-minima transforms \cite{4671118}, radial-symmetry-based voting~\cite{6698355}, supervised learning \cite{1634509} and marker clustering~\cite{YANG20142266} the watershed transform may overcome the over-segmentation problem typical for the method and can be used for segmentation of overlapping objects. A stochastic watershed is proposed in~\cite{angulo2007stochastic} to improve the performance of watershed segmentation for segmentation of complex images into few regions.
%Methods based on the watershed transform may experience difficulties with segmentation of highly overlapped objects where strong gradients are not present.
Despite all improvement, the methods based on the watershed transform may still experience difficulties with segmentation of highly overlapped objects in which a sharp gradient is not present and obtaining the correct markers is always challenging. Moreover, the watershed transform does not provide the full contours of the occluded objects.

Graph-cut is an alternative approach for segmentation of overlapping objects \cite{alkohfi,graphcut_twostage}. 
Al-Kofahi et al.~\cite{alkohfi} introduced a semi-automatic approach for detection and segmentation of cell nuclei where the detection process is performed by graph-cuts-based binarization and Laplacian-of-Gaussian filtering.  
A generalized normalized cut criterion~\cite{991040} is applied in~\cite{bernardis2010finding} for cell segmentation in microscopy images.
Softmax-flow/min-cut cutting along through Delaunay triangulation is introduced in ~\cite{chang2013invariant} for segmenting the touching nuclei.
The problem of overlapping cell separation was addressed by introducing shape's priors to the graph-cut framework in ~\cite{graphcut_twostage}.
Graph partitioning based methods become computational expensive and fail to keep the global their optimality~ \cite{6247778} when applied to the task of overlapping objects. 
Moreover, overlapping objects usually have similar image intensities in the overlapping region, which can make the graph-partition methods a less ideal approach for overlapping objects segmentation.

Several approaches have resolved the segmentation of overlapping objects within the variational framework through the use of active contours. 
The method in~\cite{5659480} incorporates a physical shape model in terms of modal analysis that combines object contours and a prior knowledge about the expected shape into an Active Shape Model (ASM) to detect the invisible boundaries of a single nucleus. The original form of ASM is restricted to single object segmentation and cannot deal with more complex scenarios of multiple occluded objects. In~\cite{acos_1} and~\cite{acos_2}, ASM was extended to address the segmentation of multiple overlapping objects simultaneously. 
Accurate initialization and computation complexity limit the efficiency of the active contour-based methods for segmentation of overlapping objects.

Morphological operations have also been applied to overlapping object segmentation. Park et al.~\cite{npa} proposed an automated morphology analysis coupled with a statistical model for contour inference to segment partially overlapping nanoparticles. Ultimate erosion modified for convex shape objects is used for particle separation, and the problem of object inference and shape classification are solved simultaneously using a Gaussian mixture model on B-splines. The method may be prone to under-segmentation with highly overlapped objects.
In~\cite{7300433}, a method for segmentation of overlapping objects with close to elliptical shape was proposed. The method consists of three steps: seed point extraction using radial symmetry, contour evidence extraction via an edge to seed point matching, and contour estimation through ellipse fitting.

Several papers have addressed the problem of overlapping objects segmentation using concave points detection~\cite{bubble,Bai20092434,Zafari2017BB, Zafari2017}. 
The concave points in the objects contour divide the contour of overlapping objects into different segments. Then the ellipse fitting is applied to separate the overlapping objects.
As these approaches strongly rely on ellipse fitting to segment object contours, they may have problems with either object's boundaries containing large scale fluctuations or objects whose shape deviate from the ellipse.

Supervised learning methods have also been applied for touching cell segmentation ~\cite{10.1007/978-3-319-24574-4_46}. 
In ~\cite{chen2013flexible}, a statistical model using PCA and distance-based non-maximum suppression is learned for nucleus detection and localization.
In~\cite{Arteta}, a generative supervised learning approach for the segmentation of overlapping objects based on the maximally stable extremal regions algorithm and a structured output support vector machine framework~\cite{hearst1998support} along dynamic programming was proposed.
The Shape Boltzmann Machine (SBM) for shape completion or missing region estimation was introduced in~\cite{Eslami2014}.
In~\cite{7952578}, a model based on the SBM was proposed for object shape modeling that represents the physical local part of the objects as a union of convex polytopes.
A Multi-Scale SBM was introduced in~\cite{YANG2018375} for shape modeling and representation that can learn the true binary distributions of the training shapes and generate more valid shapes. 
The performance of the supervised method highly depends on the amount and quality of annotated data and the learning algorithm.

Recently, convolutional neural networks (CNN)~\cite{cnn} have gained particular attention due to its success in many areas of research. 
In~\cite{cervical}
a multiscale CNN and graph-partitioning-based framework for cervical cytoplasm and nuclei detection was introduced.
In ~\cite{10.1007/978-3-319-24574-4_45} 
CNN based on Hough-voting method was proposed for cell nuclei localization. 
The method presented in~\cite{7274740} combined a CNN and iterative region merging method for nucleus segmentation and shape representation.
In~\cite{7562400}, a CNN framework is cooperated with a deformation model to segment individual cell from the overlapping clump in pap smear images.

\sloppy Although CNN based approaches have significantly contributed to the improvement of object segmentation, they come along with several issues.
CNN approaches require extensive annotated data to achieve high performance, and they do not perform well when the data volumes are small. 
Manually annotating large amount of object contour is very time-consuming and expensive. This is also the case with the nanoparticle dataset used in this study limiting the amount of data that could be used for training.
Moreover, end-to-end CNN based approaches solve the problem in a black-box manner. This is often not preferred in industrial applications as it makes it difficult to debug the method if it is not working correctly and lacks possibility to transfer or to generalize the method to different measurement setups (e.g., different microscopes for nanoparticle imaging) 
and to other applications (e.g., cell segmentation) without retraining the whole model.
%and they suffer from low interpretability of the results.  As a result, these approaches are not the ultimate solution for industrial applications where the results of the algorithm have to be translated and delivered to the general audience.
Finally, the segmentation of overlapping objects from silhouette images can be divided into separate subproblems with clear definitions. These include separating the object from the background (binarization), finding the visible part of contour for each object, and estimating the missing parts. Each of these subproblems can be solved individually making it unnecessary to utilize end-to-end models such as CNNs. 

%Therefore, we break down the problem into multiple subproblems and solve the sub-problems to obtain the final segmentation result.
%----------------------------------
\section{Overlapping object segmentation } \label{sec:method}
\begin{figure*}
\begin{center}
	{\includegraphics [width=0.9\textwidth]{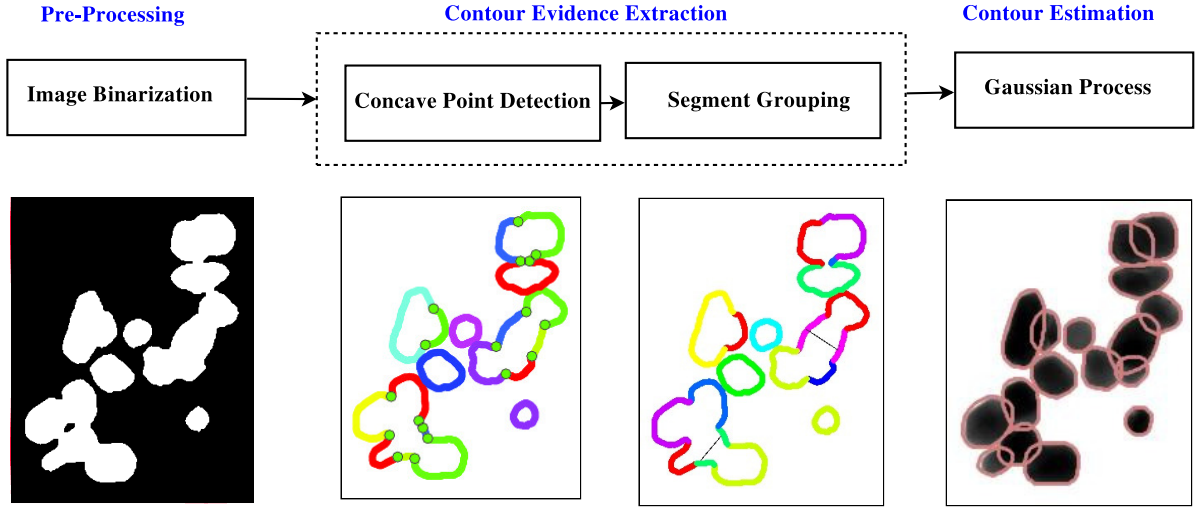}}
	\caption[]{ The workflow of the proposed segmentation method includes three major steps: pre-processing, contour evidence extraction and contour segmentation. In pre-processing, the region of the image that is covered by the object of interest is separated from the background. Contour evidence extraction infers the visible parts of the object contours and contour estimation resolves the full contours of an individual object.  }\label{fig:2}
\end{center}
\end{figure*}

\noindent
Fig.~\ref{fig:2} summarizes the proposed method.
The method is a modular framework where the performance of each module directly impacts the performance of the next module.
Given a grayscale image as an input, the segmentation process starts with pre-processing to build an image silhouette and the corresponding edge map. In our method, the binarization of the image is obtained by background suppression based on the Otsu's method~\cite{otsu1975threshold} along with the morphological opening to smooth the object boundaries. For computational reasons, the connected components are extracted and further analysis is performed for each connected component separately. The edge map is constructed using the Canny edge detector~\cite{canny}. 
However, it should be noted that the pre-processing steps are highly application specific. In this work, we consider backlit microscope images of nanoparticles where the objects of interest are clearly distinguishable from the background.
For other tasks such as cell image analysis, the binarization using the thresholding-based methods often does not provide good results due to various reasons such as noise, contrast variation, and non-uniform background.
In~\cite{Zafari2018} a machine learning framework based on the convolutional neural networks (CNN) was proposed for binarization of nanoparticle images with non-uniform background.
%
%\tuomas{If SCIA paper will be accepted we can cite it here.}
%-----------------------------------------Figure 2 (pipeline)
%----------------------------------------

The next step called contour evidence extraction involves two separate tasks: contour segmentation and segment grouping. In contour segmentation the  contour segments are recovered from a binarized image using concave points detection. 
In segment grouping, the contour segments which belong to the same objects are grouped by utilizing the branch and bound algorithm~\cite{Zafari2017BB}. 
Finally, contour evidences are utilized to estimate the full contours of the objects.
Once the contour evidence has been obtained contour estimation is carried out to infer the missing parts of the
overlapping objects.

\subsection{Concave point detection}
\noindent
Concave points detection has a pivotal role in contour segmentation and provides an important cue for further object segmentation (contour estimation).
The main goal in concave points detection (CPD) is to find the concavity locations on the objects boundaries and to utilize them to segment the contours of overlapping objects in such way that each contour segment contains edge points from one object only. 
Although a concave point has a clear mathematical definition, the task of concave points detection
from noisy digital images with limited resolution is not trivial.
Several methods have been proposed for the problem. These methods can be categorized into three groups: concave chord, skeleton and polygonal approximation based methods.

%%%%%%%%%%%%%%%%%

\subsubsection{Concave chord}
\sloppy In concave chord methods, the concave points are the deepest points on the concave region of contours that has the largest perpendicular distance from the corresponding convex hull. 
The concave regions and their corresponding convex hull chords can be obtained by two approaches as follows.
In the method proposed by Farhan \emph{et al.}~\cite{farhan2013novel}, series of lines are fitted on object contours and concave region are found as a region of contours such that the line joins the two contour points does not reside inside an object. 
The convex hull chord corresponding to that concave region is a line connecting the sixth adjacent contour point on either side of the current point on the concave region of the object contours.  
In the method proposed by Kumar \emph{et al.}~\cite{chord}, the concave points are extracted using the boundaries of concave regions and their corresponding convex hull chords.
In this way, the concave points are defined as points on the boundaries of concave regions that have maximum perpendicular distance from the convex hull chord. The boundaries and convex hull are obtained using a method proposed in~\cite{rosenfeld1985}.

\subsubsection{Skeleton}
Skeletonization is a thinning process in the binary image that reduces the objects in the image to skeletal remnant such that the essential structure of the object is preserved. 
The skeleton information along with the object's contour can also be used to detect the concave points. 
Given the extracted object contours $C$ and skeleton $SK$ 
the concave points are determined in two ways. 
In the method proposed by Samma \emph{et al.}~\cite{bnd-skeleton} the contour points and skeleton are detected by the morphological operations and then the concave points are identified as the intersections between the skeleton and contour points.
In the method proposed by Wang \emph{et al.}~\cite{skeleton}, the objects are first skeletonized and contours are extracted. 
Next, for every contour points $c_i$ the shortest distance from the skeleton is computed. 
The concave points are detected as the local minimum on the distance histogram whose value is higher than a certain threshold.

\subsubsection{Polygonal approximation}
Polygonal approximation is a well-known method for dominant point representation of the digital curves.
It is used to reduce the complexity, smooth the objects contours, and avoid detection of false concave points in CPD methods. 
The majority of the polygonal approximation methods are based on one of the following two approaches: 1) finding a subset of dominant points by local curvature estimation or 2) fitting a curve to a sequence of line segments

The methods that are based on finding a subset of dominant points by local curvature estimation work as follows.
Given the sequence of extracted contour points ${C} = \lbrace c_1,c_2,... \rbrace$, the dominants points are identified as points
with extreme local curvature. 
The curvature value $k$  for every contour points $c_i =(x_i,y_i)$ can be obtained by:
\begin{equation}
%k_{c_i} = x_i'y_i''-y_i'x_i'' / (x_i'^2+y_i'^2).
k = (x_i'y_i''-y_i'x_i'') / (x_i'^2+y_i'^2)^{3/2}.
\end{equation}

The detected dominant points $c_{d,i} \in C_{dom}$ may locate in both concave and convex regions of the objects contours. 
To detect the dominant points that are concave, various approaches exist. 
In the method proposed by Wen \emph{et al.}~\cite{curv-th} the detected dominant points are classified as concave if the value of maximum curvature is larger than a preset threshold value. 
In the method proposed by Zafari \emph{et al.}~\cite{zafari2015segmentation}, 
the dominant points are concave points if the line connecting $c_{d,i+1}$ to $ c_{d,i-1}$ does not reside inside the object.
In the method proposed by Dai \emph{et al.}~\cite{curve-area}, the dominant points are qualified as concave points if the value of the triangle area $S$ for the points $c_{d,i+1},c_{d,i}, c_{d,i-1}$ is positive and the points are arranged in order of counter-clockwise. 

The methods that are based on fitting a curve to a sequence of line segments work as follows.
First, a series of lines are fitted to the contour points.
Then, the distances from all the points of the object contour located in between the starting and ending points of the line to the fitted lines are calculated. 
The points whose distance are larger than a preset threshold value considered as the dominant point. 

After polygon approximation and dominant points detection, the concave points can be obtained using methods proposed by Bai \emph{et al.}~\cite{Bai20092434} and Zhang \emph{et al.}~\cite{bubble}.
In the method proposed by Bai \emph{et al.}~\cite{Bai20092434}, the the dominant point $c_{d,i}$ is defined as concave if the concavity value of $c_{d,i}$ is within the range of $a_1$ to $a_2$ and the line connecting $c_{d,i+1}$ and  $c_{d,i-1}$ does not pass through the inside the objects.
The concavity value of $c_{d,i}$ is defined as the angle between lines $(c_{d,i-1} , c_{d,i} )$ and $(c_{d,i+1} , c_{d,i})$ as follows:

\begin{align}
%\hspace*{1.5cm}
\begin{dcases}
\displaystyle 
\mid 
\gamma_1 - \gamma_2\mid & \hspace{5pt}\text{if}  \ |\gamma_1 - \gamma_2| < \pi\\
\vspace*{18cm}
\pi  - |\gamma_1 - \gamma_2| & \hspace{-1pt} \text       {otherwise}, 
\end{dcases}
\end{align}

\noindent
where, $\gamma_1$ is the angle of the line $(c_{d,i-1} , c_{d,i} )$ and $\gamma_2$ is the angle of the line $(c_{d,i+1} , c_{d,i})$ defined as:
\begin{align*}
\gamma_1 = \tan^{-1} ((y_{d,i-1} - y_{d,i} )/(x_{d,i-1} - x_{d,i} )),  \\
\gamma_2 = \tan ^{-1} ((y_{d,i+1} - y_{d,i} )/(x_{d,i+1} - x_{d,i} )).
\end{align*}

In the method proposed by Zhang \emph{et al.}~\cite{bubble}, the dominant point $c_{d,i} \in C_{dom}$ is considered to be a concave point if $\vv{c_{d,i-1}c_{d,i}}\times \vv{c_{d,i}c_{d,i+1}}$ is positive:
\begin{equation} \label{eq:3:concavepoint}
\begin{split}
{C}_{con} = \lbrace c_{d,i} \in C_{dom} \ : \ \vv{c_{d,i-1}c_{d,i}}\times \vv{c_{d,i}c_{d,i+1}}>0 \rbrace
\end{split}.
\end{equation}

\subsection{Proposed method for concave point detection}
Most existent concave point detection methods require a threshold value that is selected heuristically ~\cite{bubble,Bai20092434}. 
In fact, finding the optimal threshold value is very challenging, and a unique value may not be suitable for all images in a dataset.

We propose a parameter free concave point detection method that relies on polygonal approximation by fitting the curve with a sequence of line segments.
Given the sequence of extracted contour points ${C} = \lbrace c_1,c_2,... \rbrace$, first, the dominant points are determined by co-linear suppression.
To be specific, every contour point $c_i$ is examined for co-linearity, while it is compared to the previous and the next successive contour points. 
The point $c_i$ is considered as the dominant point if it is not located on the line connecting $c_{i-1}
=(x_{i-1},y_{i-1})$ and $c_{i+1}=(x_{i+1},y_{i+1})$
and the distance $d_i$ from $c_{i}$ to the line connecting $c_{i-1}$ to $c_{i+1}$ is bigger than a pre-set threshold $d_i> d_{th}$.

\begin{figure}[h!]
\begin{center}
	{{\includegraphics [width=.3\textwidth]{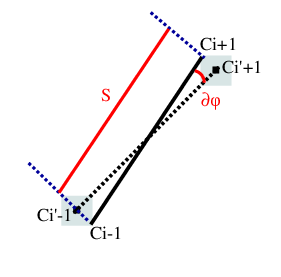}}}
\end{center}
\caption[]{Representation of the actual and the digitized lines.  }\label{fig:4.2}
\end{figure}

The value of $d_{th}$ is selected automatically using the characteristics of the line as in~\cite{prasad2012polygonal}.
In this method, first the line connecting $c_{i-1}$ to $c_{i+1}$ is
digitized by rounding the value  $(x_{i-1},y_{i-1})$ and $(x_{i+1},y_{i+1})$ to their nearest integer. 
Next, the angular difference between the numeric tangent of line $c_{i-1}c_{i+1}$ and the digital tangent of line $c'_{i-1}c'_{i+1}$ is computed as (see Fig. \ref{fig:4.2}):

\begin{equation}\label{eq:3:cb_dist}
\begin{split}
\delta\phi = \tan^{-1}(m)-\tan^{-1}(m'),
\end{split}
\end{equation}
where, $m$ and $m'$ are slop of actual and digital line respectively. Then, value of $d_{th}$ is given by:
\begin{equation}\label{eq:3:cb_dist}
d_{th} = S\delta\phi,
\end{equation}
where $S$ is length of actual line.

Finally, the detected dominant points are qualified as concave if they met the criterion by Zhang (Eq.~\ref{eq:3:concavepoint}) for concave point detection.
We referred the proposed concave point detection as Prasad+Zhang~\cite{Zafari2017}.

\subsection{Segment grouping}\label{sec:seggrp}
\noindent
Due to the overlap between the objects and the irregularities in the object shapes, a single object may produce multiple contour segments. Segment grouping is needed to merge all the contour segments belonging to the same object.

Let $S= \lbrace S_1,S_2,...,S_N \rbrace$ be an ordered set of $N$ contour segments in a connected component of an image.
The aim is to group the contour segments into $M$ subsets such that the contour segments that belong to individual objects are grouped together and $M \leq N$.
Let $\omega_i$ be the group membership indicator giving the group index to which the contour segment $S_i$ belongs to.
Denote $\Omega$ be the ordered set of all membership indicators: $\lbrace \omega_1,\omega_2,....,\omega_N \rbrace$.

The grouping criterion is given by a scalar function $J(\cdot{})$ of $\Omega$ which maps a possible grouping of the given contour segments onto the set of real numbers $\mathbb{R}$. 
That is, $J$ is the cost of grouping that ideally measures how the grouping $\Omega$ resembles the true contour segments of the objects.
The grouping problem for the given set of $S$ is to find the optimal membership set $\Omega^{*}$ such that the grouping criterion (cost of grouping) is minimized:
\begin{equation}\label{eq:groupingproblem}
\Omega^{*} = \argmin_{\Omega} \ J(\Omega;S).
\end{equation}

\sloppy We formulated the grouping task as a combinatorial optimization problem and solved using the branch and bound (BB) algorithm~\cite{Zafari2017BB}. 
%The efficiency of this algorithm depends upon the definition of the grouping criterion. 
The BB algorithm can be applied with any grouping criterion $J$. However, the selection of the grouping criterion 
%has significant effect on 
significantly affects 
the overall method performance. In this work the grouping criterion is a hybrid cost function consisting of two parts: 1) \textit{generic part} ($J_{\mathrm{concavity}}$) that encapsulate the general convexity properties of the objects and 2) \textit{specific part} that encapsulates the properties of objects that are exclusive to a certain application, e.g., symmetry ($J_{\mathrm{symmetry}}$) and ellipticity ($J_{\mathrm{ellipticity}}$). The cost function is defined as follows:
\begin{equation}
J = \underbrace{ J_{\mathrm{concavity}}}_\text{Generic}\underbrace{+\alpha J_{\mathrm{ellipticity}}+ {\beta J_\mathrm{symmetry}}}_\text{Specific},
\end{equation}
where $\alpha$, $\beta$  are the weighting parameters. 
The generic part encourages the convexity assumption of the objects in order to penalize the grouping of the contour segments belonging to different objects. 
This is achieved by incorporating a quantitative concavity measure. Given two contour segments $S_i$ and $S_j$ the generic part of the cost function is defined as follows:
\begin{align}
\small
J_{\mathrm{concavity}} = 
\begin{cases}
1, & $  concave point between 
$S_i$ and $S_j$ $\\ 
(1-\frac{A_{S_i \cup S_j}}{A_{\mathrm{ch},S_i \cup S_j}}) & \  \text{otherwise}, 
\end{cases} &
\end{align}
\noindent
where ${A_{S_i \cup S_j}}$ is the area of a region bounded by $S_i$, $S_j$, and the lines connecting the endpoint of $S_j$ to the starting point of $S_i$ and the end point of $S_i$ to the starting point of $S_j$.
$A_{\mathrm{ch},S_i \cup S_j}$ 
%is the corresponding area of the convex hull bounded by $S_i$ and $S_j$.
is the upper bound for area of any convex hull with boundary points $S_i$ and $S_j$.
The specific part is adapted to consider the application criteria and certain object properties. Considering the object under examination, several functions can be utilized. 

Here, the specific term is formed by the ellipticity and symmetry cost functions.
% \textit{ellipticity:}
The ellipticity term is defined by measuring the discrepancy between the fitted ellipse and the contour segments~\cite{bubble}. 
Given the contour segment $S_i$ consisting of $n$ points, $S_i = \lbrace (x_{k},y_{k}) \rbrace_{k=1}^{n}$, and the corresponding fitted ellipse points, $S_{f,i} =\lbrace (x_{f,k},y_{f,k}) \rbrace_{k=1}^{n}$, the ellipticity term is defined as follows:
\begin{equation}
\begin{split}\label{eq:bubble.add.1}
\mathrm{J_{ellipticity}} = \frac{1}{n} \mathlarger{\sum}_{k=1}^n {\sqrt{(x_k - x_{f,k})^2 +( y_k-y_{f,k})^2}}.
\end{split}
\end{equation}

The \textit{symmetry} term penalizes the resulting objects that are non-symmetry.
Let $o_i$ and $o_j$ be the center of symmetry of the contour segments $S_i$ and $S_j$ obtained by aggregating the normal vector of the contour segments. 
The procedure is similar to fast radial symmetry transform~\cite{frs}, but the gradient vectors are replaced by the contour segments' normal vector. 
This transform is referred to as \textit{normal symmetry transform} (NST).

In NST, every contour segment point gives a vote for the plausible radial symmetry at some specific distance from that point. 
Given the distance value $n$ of the predefined range $[{R}_{min} \ {R}_{max}]$,  for every contour segment point $(x,y)$, NST determines the negatively affected pixels ${P}_{-e}$ by 
\begin{equation}\label{eq:3:pneffectedpixel}
\begin{split}
&{P}_{-ve}(x,y) = (x,y) - \mathrm{round} \mathlarger{\mathlarger{(}} \frac{\ve{n}(x,y)}{\Vert\ve{n}(x,y)\Vert} n \mathlarger{\mathlarger{)}},\\
\end{split}
\end{equation} 
and increment the corresponding point in the orientation projection image $\ve{O}_n$  by  $1$.
The symmetry contribution $\ve{S}_n$ for the radius $n\in [{R}_{min},{R}_{max}]$ is formulated as
\begin{equation}\label{eq:3:fm}
\begin{split}
\ve{S}_n(x,y) = {\mathlarger{\mathlarger{(}}\frac{\vert \tilde{\ve{O}}_n(x,y) \vert}{k_n}\mathlarger{\mathlarger{)}}}^{{}}
\end{split},
\end{equation}
where $k_n$ is the scaling factor that normalizes $\ve{O}_n$ across different radii. $\tilde{\ve{O}}_n$ is defined as
\begin{equation}
\begin{split}
\tilde{\ve{O}}_n(x,y)=\begin{cases}
\ve{O}_n(x,y), & \text{if } \ve{O}_n(x,y)<k_n,\\
k_n, & \text{otherwise}.
\end{cases}
\end{split}
\end{equation}
The full NST transform $\ve{S}$, by which the interest symmetric regions are defined, is given by the average of the symmetry contributions over all the radii $n \in [{R}_{min},{R}_{max}]$ considered as
\begin{equation}\label{eq:3:s}
\ve{S} =  \frac{1}{\vert N \vert} \sum_{n \in [{R}_{min},{R}_{max}]} \ve{S}_n.
\end{equation}
 The center of symmetry $o_i$ and $o_j$  of the contour segments are estimated as the average locations
of the detected symmetric regions in $\ve{S}$. The symmetry term $J_{\mathrm{symmetry}}$ is defined as the Euclidean distance between $o_i$ and $o_j$ as
\begin{equation}
\begin{split}\label{eq:bubble.add.1}
\mathrm{J_{symmetry}} = \vert o_i - o_j \vert.
\end{split}
\end{equation}
The distance is normalized to [0,1] according to the maximum diameter of the object.

\subsection{Contour estimation}\label{sec:step3} 
\noindent
\sloppy Segment grouping results in, a set of contour points for each object. Due to the overlaps between objects, the contours are not complete and the full contours need to be estimated. 
Several methods exist to address the contour estimation problem.
A widely known approach is Least Square Ellipse Fitting (LESF)~\cite{fitzgibbon1999direct}. 
The purpose of LESF is to computing the ellipse parameters by minimizing the sum of squared algebraic distances from the known contour evidences points to the ellipse. 
The method is highly robust and easy to implement, however, it fails to accurately approximate complicated non-elliptical shapes.

Closed B-splines is another group of methods for the contour estimation. 
In~\cite{npa}, the B-Splines with Expectation Maximization (BS) algorithm was used to infer the full contours of predefined possible shapes. 
The reference shape and the parameters of the B-spline of the object are comprised in the Gaussian mixture model. The inference of the curve parameters, namely shape properties and the parameters of B-spline, from this probabilistic model is addressed through the Expectation Conditional Maximization (ECM) algorithm \cite{meng1993ecm}. 
The accuracy of the method depends on the size and diversity of the set of reference shapes and gets low if the objects highly overlap.

\subsection{Proposed method for contour estimation}
To address the issues with the existing methods, we propose a Gaussian process based method for the contour estimation.
Gaussian processes (GP) are nonparametric kernel-based probabilistic models that have been used in several machine learning applications such as regression, prediction, and contour estimation~\cite{martin2005use,RICHARDSON2017209}.
In ~\cite{YANG2018387} a GP model based on charging curve is proposed for
state-of-health (SOH) estimation of lithium-ion battery.
In~\cite{pmlr-v28-wilson13} GP is applied for pattern discovery and extrapolation.
In ~\cite{7274766} a regression method based on the GP is proposed for prediction of wind power.
%
%\tuomas{It is difficult to see how the above three references are relevant for this application.}

The primary assumption in GP is that the data can be represented as a sample from a multivariate Gaussian distribution. 
GP can probabilistically estimate the unobserved data point, $f^*$, based on the observed data, $f$, assuming that the both observed data point and unobserved data can be represented as a
sample from a multivariate Gaussian distribution
\begin{equation}\label{eq:gp}
\begin{pmatrix}
f \\
f^* \\ 
\end{pmatrix}
= \mathcal{N}
\begin{pmatrix}
\begin{pmatrix}
\mu_{\mathbf{x}} \\
\mu_{\mathbf{x}^*} \\ 
\end{pmatrix}
,
\begin{pmatrix}
\ve{K}_\mathbf{x\mathbf{x}}& \ve{k}_\mathbf{x^*}^T \\
\ve{k}_\mathbf{x^*}  & \ve{K}_\mathbf{x^*\mathbf{x^*}} \\ 
\end{pmatrix}
\end{pmatrix},
\end{equation}
where $\mathbf{K}$ is the $n \times n$ covariance matrix and $\ve {k}_{\mathbf{x}^*}$ is an
$n \times 1$ vector containing the covariances between $x^*$ and $ x_1,x_2,\ldots, x_n$  given by
\begin{equation} \label{eq:21}
[\mathbf{K}]_{ij} = \kappa(x_i, x_j),
\end{equation}
and
\begin{equation} \label{eq:22}
\mathbf{k}_{\mathbf{x}^*} = [ \kappa(x_1, x^*), \kappa(x_2, x^*),\ldots, \kappa(x_n, x^*) ]^T.
\end{equation}
%%%%%%%%%%%%%%%%%%%%%%%%%%%%%%%%%%%%%%%%%%%%%%%%%%%%%%%%%
$\kappa(.,.)$  is defined by the Matern covariance function~\cite{rasmussen2006gaussian} as follows:
\begin{equation}\label{eq:matern}
\begin{split}
\kappa(x_i, x_j) = \frac{2^{1-v}}{\Gamma(v)} (\frac{\sqrt{2v}|d|}{l})^ v K_v(\frac{\sqrt{2v}|d|}{l})
\end{split},
\end{equation}
where $x_i$, $x_j$, $i \neq j$, $i = 1, 2, \ldots, n$ are a pair of data point, $l$ is the characteristic length scale, $d$ is the Euclidean distance between the points $x_i$ and $x_j$ and $K_v $ is the modified Bessel function of order $v$. 
The most common choices for $v$ are $v = 3/2$ and $v = 5/2$ that are denoted by Matern 3/2 and Matern 5/2. 

Let $D~=~\{(x_i,y_i)\}_{i=1}^{n}$  be the input data, $\mathbf{x}=[x_1, \ldots, x_n]^T$ be the input vector and $\mathbf{y} =[y_1, \ldots, y_n]^T$ be the corresponding observations vector given by a latent function $f(.)$.  
GP assumes the latent function comes from a zero-mean Gaussian and the observations are modeled by 
\begin{equation}\label{eq:model}
\begin{split}
{y_i} =  f({x_i}) + \epsilon_i, \quad
\epsilon_i \sim \mathcal{N}(0, \sigma^2).
\end{split}
\end{equation}
Given the $D, f$ and  a new point $\mathbf{x^*}$, the GP
defines the posterior distribution of the latent function $f(.)$ at the new point $f(\mathbf{x}^*)$ as Gaussian and obtains the best estimate for $\mathbf{x^*}$ by
the mean and variance of the posterior distribution as follows:
\begin{equation}\label{eq:18}
\begin{split}
p(f^* |\mathbf{x}^*,{D}) = \mathcal{N} ({f^*} | 
{\mu_{\mathbf{x}^*}}, {\sigma^2_{\mathbf{x}^*}} ),
\end{split}
\end{equation}
where
\begin{equation}\label{eq:19}
\mathbf{\mu}_\mathbf{x^*} = \mathbf{k}_{\mathbf{x}^*}^T\mathbf{K}_{\mathbf{x}\mathbf{x}} ^{-1} \mathbf{y},
\end{equation}
and
\begin{equation} \label{eq:20}
\mathbf{\sigma}_\mathbf{x^*}^2 = \mathbf{K}_{\mathbf{x^*}\mathbf{x^*}} - \mathbf{k}_{\mathbf{x}^*}^T \mathbf{K}_{\mathbf{x}\mathbf{x}}  ^ {-1}\mathbf{k}_{\mathbf{x}^*}.
\end{equation}

To exploit GP for the task of contour estimation and to limit the interpolation interval to a finite domain, the polar representation of contour evidence is needed. 
Polar representation has three advantages.
First, it limits the interpolation interval to a finite domain of $[0,2\pi]$, second, it exploits the correlation between the radial and angular component, and third, it reduces the number of dimensions compared to GP based on parametric Cartesian representation.

In \emph{GP with polar form (GP-PF)}, the contour evidence points are represented as a function of the radial and angular components.
Given the contour evidence points $x$ and $y$, their polar transform is defined as follows:
\begin{equation}\label{eq:polar}
\begin{split}
{\theta} &= \arctan{\left( \frac{y}{x}\right)}, \\
r &= \sqrt{x^2 + y^2}.
\end{split}
\end{equation}
Here, the angular component $\theta$ is considered to be the input to the GP and the radial $r$ as the corresponding observations described by the latent function $r = f(\theta)$. 
To enforce the interpolation to a closed periodic curve estimation such that $f(\theta) = f(\theta + 2\pi) = r$, 
the input set is defined over the finite domain $\theta^* \in [0, 2\pi]$ as $D = \{(\theta_i,r_i)\}_{i=1}^{n} \cup  \{(\theta_i+2\pi,r_i), ...\}_{i=1}^{n}$. 
The best estimate for $r^*$ is obtained by the mean of the distribution
%-----------------------------
\begin{equation}\label{eq:mean}
\mathbf{\mu}_{\theta^*} = \mathbf{k}_{{\theta}^*}^T\mathbf{K}_{{\theta}{\theta}} ^{-1} \ve{r}.
\end{equation}
The GP-PF method is summarized in Algorithm \ref{alg:SPGP-PF}\vspace{1cm}.
%%%%%%%%%%%%%%%%%%%%%%%%%%%%%%%%%%%%%%%%%%%%%

{
\RestyleAlgo{algoruled}\SetAlgoVlined\LinesNotNumbered
\newcommand{\KwParameters}[1]{\textbf{Parameters: }{#1}} 
\begin{algorithm}
	\DontPrintSemicolon
	\KwIn{Contour evidence coordinates $x$, $y$.}
	\KwOut{Estimated objects' full contour points $x^*$, $y^*$.}
	\KwParameters{Covariance function $k$}
	\begin{enumerate}\itemsep0em
		\item Perform mean centering for $x$, $y$ by subtracting their mean from all their values.
		\item Retrieve polar coordinates $\theta$, $r$ from $x$, $y$ using Eq.~[\ref{eq:polar}].
		\item Define the input set and the corresponding output for the observed contour points as \\ $D = \{(\theta_i,r_i)\}_{i=1}^{n} \cup  \{(\theta_i+2\pi,r_i), ...\}_{i=1}^{n}$.
		\item Define the new input set for unobserved contour points by $ \theta^* \in [0, 2\pi]$.
		\item Estimate the covariance function, $K_{\theta^*\theta^*} , k_{\theta^*}, K_{\theta\theta}$ using Eq.~[\ref{eq:matern}].
		\item Predict the output, $r^*$, on the new input set, $\theta^*$, using the predictive mean using Eq.~[\ref{eq:mean}].             
		\item Transform $\theta^*$, $r^*$ from polar to Cartesian coordinates $x^*$, $y^*$.
	\end{enumerate}
	\caption{Contour Estimation by GP-PF}
	\label{alg:SPGP-PF}
\end{algorithm}
}

\section{Experiments}\label{sec:exp}

\noindent
This section presents the data, the performance measures, and the results for the concave points detection, contour estimation, and segmentation.

\subsection{Data}

\noindent
The experiments were carried out using one synthetically generated dataset and one dataset from a real-world application.

\begin{figure*} [ht!]
\begin{center}
	\begin{tabular}{c c }
		\frame{\includegraphics [width=.45\textwidth,height=0.25\textheight]{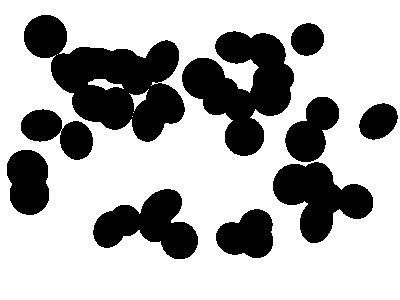}}&
		\frame{\includegraphics [width=.45\textwidth,height=0.25\textheight]{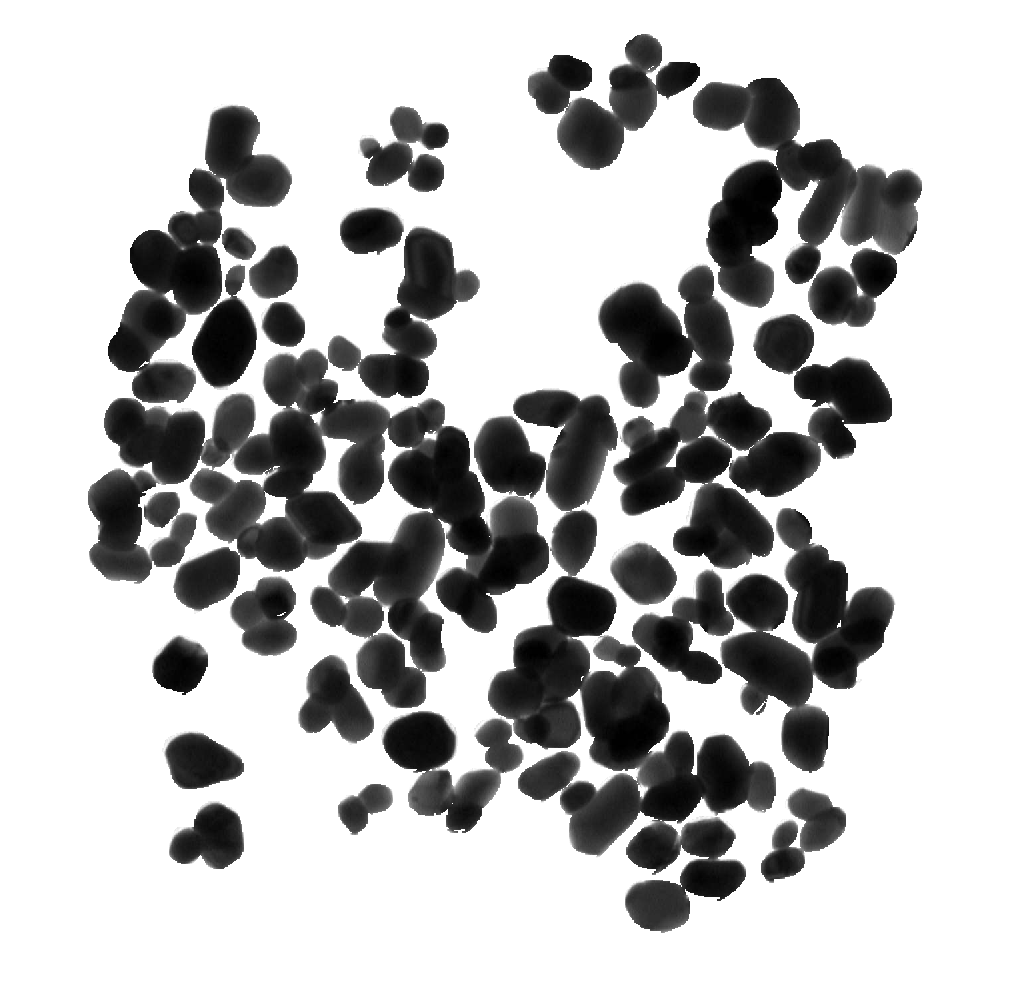}} \\
		(a)&(b)
	\end{tabular}
	\caption{ Example images from the datasets studied: (a) Synthetic dataset with the maximum overlap of 40\%; (b) Nanoparticles dataset.  } \label{fig:3}
\end{center}
\end{figure*}

The synthetic dataset (see Fig.~\ref{fig:3}(a)) consists of images with overlapping ellipse-shape objects that are uniformly randomly scaled, rotated, and translated. Three subsets of images were generated to represent different degrees of overlap between objects. The dataset consists of 150 sample images divided into three classes of overlap degree. The maximum rates of the overlapping area allowed between two objects are 40\%, 50\%, and 60\%, respectively, for the first, second, and third subset. Each subset of images in the dataset contains 50 images of 40 objects. The minimum and maximum width and length of the ellipses are 30, and 45 pixels. The image size is $300 \times 400$ pixels. 

The real dataset (nanoparticles dataset) contains nanoparticles images captured using transmission electron microscopy  (see Fig.~\ref{fig:3}(b)). In total, the dataset contains 11 images of $4008 \times 2672$ pixels. Around 200  particles were marked manually in each image by an expert. In total the dataset contains 2200 particles.The annotations consist of manually drawn contours of the particles. This information was also used to determine the ground truth for concave points. Since not all the particles were marked, a pre-processing step was applied to eliminate the unmarked particles from the images. It should be noted that the images consist of dark objects on a white background and, therefore, pixels outside the marked objects can be colored white without making the images considerably easier to analyze.
%------------------------------ Figure 3 (example images of datasets)

%--------------------------------------------
\subsection{Performance Measures} \label{ssec:permer}

\noindent
To evaluate the method performance and to compare the methods, 
the following performance measures were used: 
The detection performance for both concave points and objects were measured
using True Positive Rate (TPR), Positive Predictive Value (PPV), and Accuracy 
(ACC) defined as follows:
\begin{align}
TPR &= \frac{TP}{TP+FN}, \\
PPV &= \frac{TP}{TP+FP}, \\
ACC &= \frac{TP}{TP+FP+FN},
\end{align}
where True Positive (TP) is the number of correctly detected
concave points or segmented objects, False Positive (FP) is the number 
of incorrectly detected concave points or segmentation results, and 
False Negative (FN) is the number of missed concave points or objects.
To determine whether a concave point was correctly detected (TP), the distance to the ground truth concave points was computed and the decision
was made using a predefined threshold value. The threshold value was set 
to 10 pixels. Moreover, the average distance (AD) from detected concave 
points to the ground truth concave points was measured for the concave 
points detection.
To decide whether the segmentation result was correct or incorrect, 
Jaccard Similarity coefficient (JSC)~\cite{Choi10asurvey} was used. 
Given a binary map of the segmented object $O_s$ and the ground truth 
particle $O_g$, JSC is computed as
\begin{align}
\hspace{1cm} & JSC = \frac{O_s \cap O_g}{O_s \cup O_g}. &
\end{align}
%-----------------------------------------------------
The threshold value for the ratio of overlap (JSC threshold) 
was set to 0.6. The average JSC (AJSC) value was also used as a 
performance measure to evaluate the segmentation performance.

\subsection{Parameter selection}
\noindent
The proposed method requires the following major parameters.

\sloppy{\textit{Weighting parameters $\alpha$, and $\beta $}} define to what extent the ellipticity and symmetry features must be weighted in the computation of the grouping cost $J$. 
A lower value of $\alpha$ and $\beta $ emphasizes non-ellipticity and non-symmetry features where higher values ensure that objects have perfect elliptical symmetry shapes. 
This parameter is useful when the objects are not perfectly elliptical.
In this work, the weighting parameters $\alpha$, and $\beta $ were set to $0.1$ and $0.9$ 
to obtain the highest performance of segment grouping.

{\textit{Type of the covariance function}} determines how the covariance
function measure the similarity of any two points of the data, and specifies the desired form of the output function and affects the final contour estimation result. In this work, the contour estimation is performed using the \textit{GP-PF} method with the Matern 5/2
covariance function as a more flexible and general one.

The parameters of the concave points detection methods were set experimentally to obtain the best possible result for each method.

%----------------------------------------
\subsection{Results} 

\subsubsection{Concave points detection} \label{concavepoint extraction}

\noindent
\sloppy The results of the concave points detection methods applied to real nanoparticles and synthetic datasets are presented in Tables~\ref{tab:2} and~\ref{tab:1}, respectively. 
The results have been reported with the optimal parameters setup for each listed methods. 
The chosen nanoparticle dataset is challenging for CPD experiments 
since it contains objects with noisy boundaries. 
Here, the noise includes the small or large scale fluctuations on the object boundaries 
that can affect the performance of the concave points detection methods.
The results on the nanoparticles dataset (Table~\ref{tab:2}) show 
that the Prasad+Zhang method outperforms the others with the highest TPR and ACC values and the competitive PPV and AD values.
From the results with the synthetic dataset (Table~\ref{tab:1}) 
it can be seen that Prasad+Zhang achieved the highest value of ACC. Zhang~\cite{bubble} scored the highest values of TPR and AD, but it suffers from the low values of PPV and ADD. 
This is because the method produces a large number of false concave points.
Comparing the Zhang~\cite{bubble} and Prasad+Zhang~\cite{Zafari2017} CPD methods shows the advantage of the proposed parameter selecting strategy in concave points detection 
that makes the Prasad+Zhang method a more robust in presence of objects with noisy boundaries.
In terms of PPV, Zafari~\cite{zafari2015segmentation} and Kumar~\cite{concavity} ranked the highest. 
However, these approaches tend to underestimate the number of concave points as they often fail to detect the concave points at a low depth of concave regions. 
Considering all the scores together, the results obtained from the synthetic dataset showed 
that the Prasad+Zhang method achieved the best performance.

\begin{figure*} [h!]
\begin{center}
	\begin{tabular}{c c c}
		{\includegraphics [width=.3\textwidth]{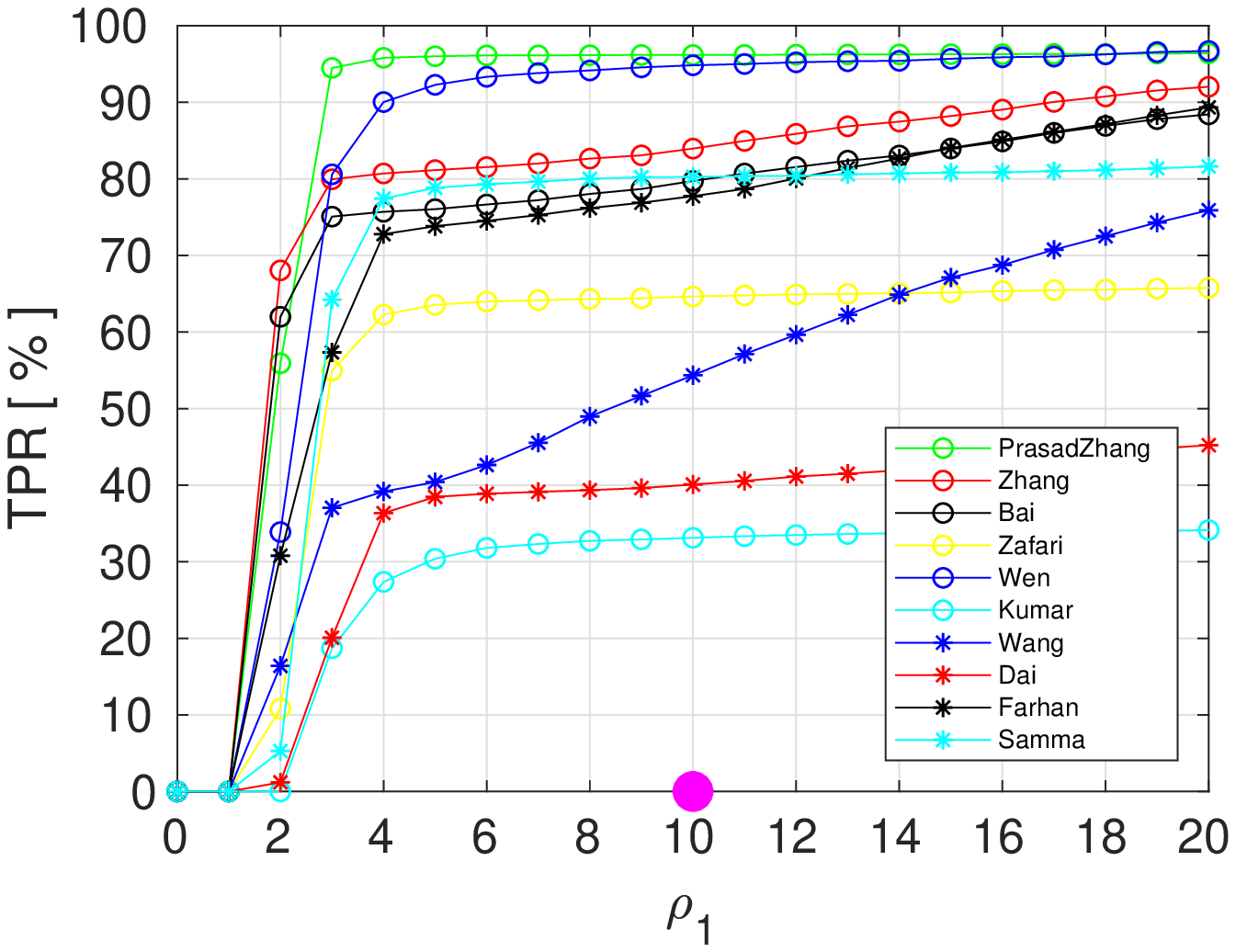}}&
		{\includegraphics [width=.3\textwidth]{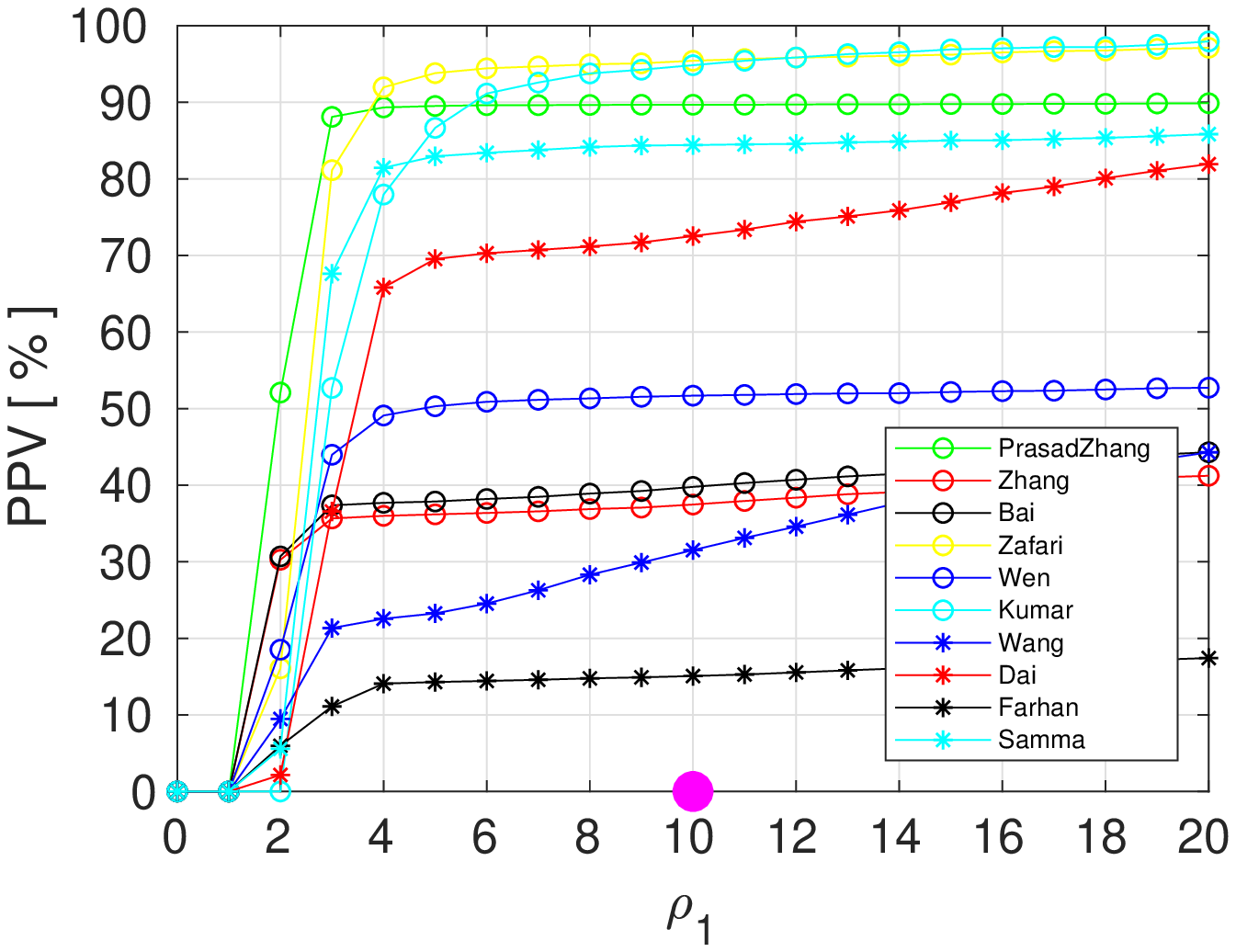}}&
		{\includegraphics [width=.3\textwidth]{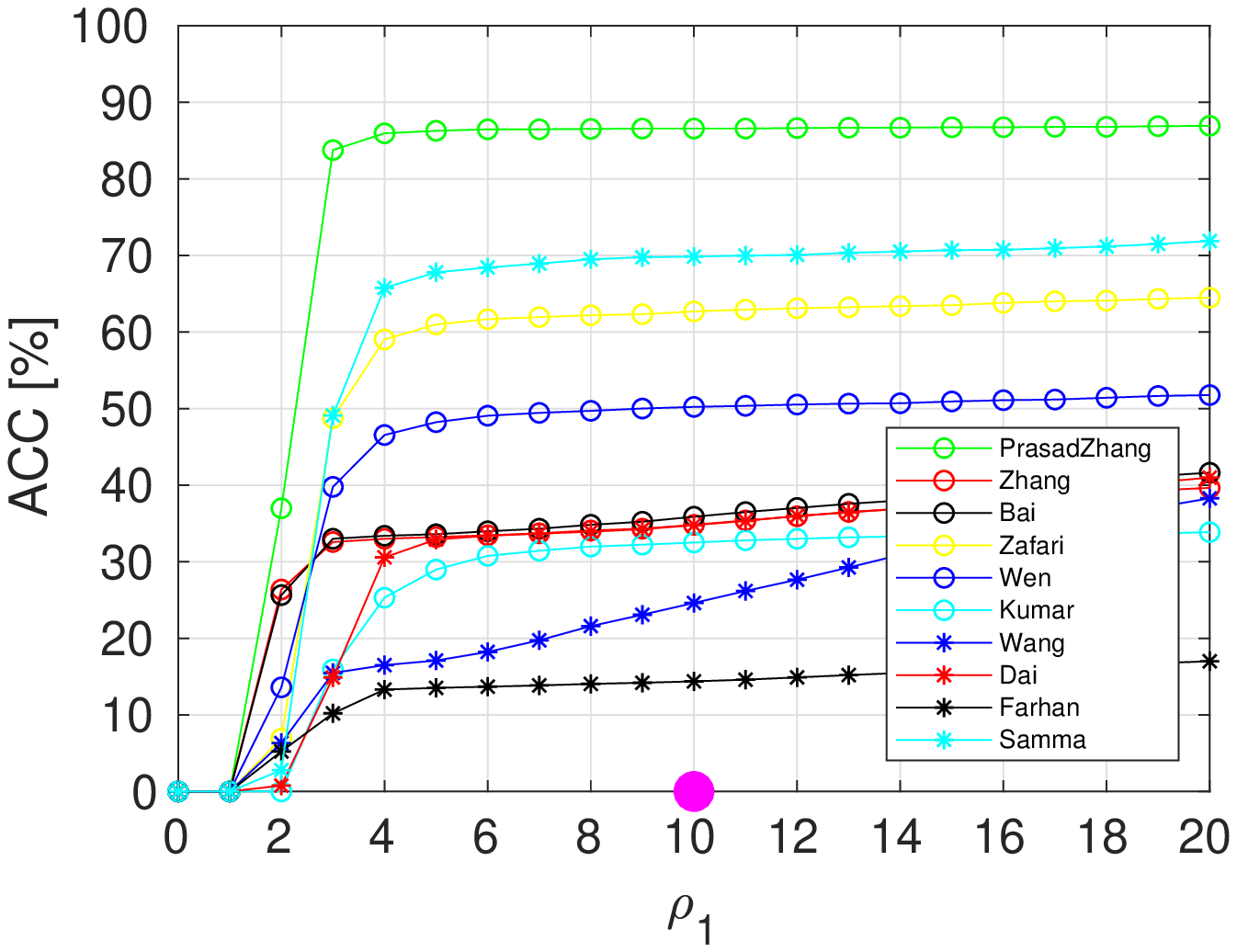}} \\
		(a)&(b)&(c)
	\end{tabular}
	\caption{Performance of concave detection methods with different values of the distance threshold ($\rho_1$) in the nanoparticles dataset: (a) TPR; (b) PPV; (c) ACC.  } \label{fig:6.8}
\end{center}
\end{figure*}

The results presented in Tables~\ref{tab:2} and~\ref{tab:1} were obtained by a fixed distance threshold value of 10 pixels.
To check the stability of the results with respect to distance threshold ($\rho_1$), the distance threshold was examined by the nanoparticles datasets. 
The effect of $\rho_1$ on the TPR, PPV and ACC scores captured with the concave point detection methods are presented in Fig.~\ref{fig:6.8}.
As it can be seen,  the Prasad+Zhang method achieved the highest score of TPR and ACC regardless of the selected threshold.   
It is worth mentioning that at low values of $\rho_1$, the Prasad+Zhang method outperforms all methods with respect to PPV. 
This is clearly since the Prasad+Zhang method can detect the concave points more precisely.

%--------------------------------------------Table 2
\begin{table}[h]
\centering
\begin{tabular}{p{2.2cm}>
		{\centering\arraybackslash}p{0.4cm}>
		{\centering\arraybackslash}p{0.4cm}>
		{\centering\arraybackslash}p{0.4cm}>
		{\centering\arraybackslash}p{0.4cm}>
		{\centering\arraybackslash}p{0.4cm}>
		{\centering\arraybackslash}p{0.4cm}}
	%  \toprule
	\toprule
	\multirow{2}{*}{Methods}&   TPR & PPV  & ACC &AD & Time &\\
	&[\%] &[\%]&[\%]&[pixel]&(s)\\
	\midrule
	Prasad+Zhang~\cite{Zafari2017} &  \textbf{96}  & {90} & \textbf{87}  &1.47 & \textbf{0.50} \\    
	Zhang \cite{bubble}      &  84 & 38 & 35   & \textbf{1.42} & 2.28\\
	Bai \cite{Bai20092434}       &    79 & 40& 36   & 1.55 &2.52 \\
	Zafari \cite{zafari2015segmentation}    &   65& \textbf{95} & 64&2.30 & 2.48\\
	Wen \cite{curv-th}    &    94& 51 &50 &1.97 & 2.13\\
	Dai  \cite{curve-area}  &     54        & 87 & 50& 2.84  & 2.50\\
	Samma  \cite{bnd-skeleton}       & 80   & 84 & 70 &2.28  & 2.95\\
	Wang \cite{skeleton}     &   65 & 58 & 44&2.37  & 4.95\\
	Kumar \cite{chord}   &  33    & 94 &  31   &3.09   & 8.17\\
	Farhan~\cite{farhan2013novel} &  77     & 15 & 14 & 2.09& 8.82\\
	
	\bottomrule
\end{tabular}
\caption{Comparison of the performance of concave points detection methods on the nanoparticles dataset.}\label{tab:2}
\end{table}

%--------------------------------------------- Table 1(comparison of CPD)
\begin{table}[h]
\centering
\begin{tabular}{p{2.2cm}>
		{\centering\arraybackslash}p{1.5cm}>
		{\centering\arraybackslash}p{0.4cm}>
		{\centering\arraybackslash}p{0.5cm}>
		{\centering\arraybackslash}p{0.5cm}>
		{\centering\arraybackslash}p{0.7cm}>
		{\centering\arraybackslash}p{0.5cm}}
	%\vspace{10pt}
	%\toprule
	\midrule
	\multirow{2}{*}{Methods}& Overlapping& TPR & PPV & ACC & AD \\
	&[\%]&[\%]&[\%]&[\%]&[pixel]\\
	\midrule
	Prasad+Zhang~\cite{Zafari2017} & 40& 97  & {97} &\textbf{94} &1.47       \\ 
	Zhang~\cite{bubble}&40& \textbf{ 98} & 37 &37& \textbf{1.25} \\
	Bai~\cite{Bai20092434} &40& 92 & 46 &44& 1.34 \\
	Zafari~\cite{zafari2015segmentation}  &  40 & 81        &    \textbf{98} &80 &2.04 \\
	Wen~\cite{curv-th}     &  40 & 97       & 61 &60& 1.75 \\
	Dai~\cite{curve-area}  &  40 & 71       & 97 &70& 2.58  \\
	Samma~\cite{bnd-skeleton} &  40 & 75    & 83 &65& 2.20  \\
	Wang~\cite{skeleton}    &  40 & 87      & 86 &76& 1.87  \\
	Kumar~\cite{chord} & 40 & 64    & \textbf{98} &62& 2.64  \\
	Farhan~\cite{farhan2013novel}      &  40 & 91   & 18 &18& 1.77 \\
	
	\midrule
	Prasad+Zhang~\cite{Zafari2017} & 50&  \textbf{97} & 96 &\textbf{93}& 1.50 \\
	Zhang~\cite{bubble}  & 50&  96 & 38 &38& \textbf{1.25 } \\
	Bai~\cite{Bai20092434}   & 50& 86 & 47&44& 1.34  \\
	Zafari~\cite{zafari2015segmentation}  &  50 & 77        & \textbf{98} &76& 2.08  \\
	Wen~\cite{curv-th}     &  50 & 94       & 62 &60& 1.81  \\
	Dai~\cite{curve-area}  &  50 & 68       & 97 &66& 2.57  \\
	Samma~\cite{bnd-skeleton}  &  50 & 73   & 83 &63& 2.24 \\
	Wang~\cite{skeleton}     &  50 & 83     & 86 &74& 1.95  \\
	Kumar~\cite{chord} &  50 & 60   &\textbf{98}  &59& 2.62  \\
	Farhan~\cite{farhan2013novel}       &  50 & 87  & 19 &18& 1.81  \\
	\midrule
	Prasad+Zhang~\cite{Zafari2017} &  60&  94 & 96 &\textbf{91}& 1.51\\
	Zhang~\cite{bubble} &  60 &  \textbf{96} & 40 &39& \textbf{1.31}  \\
	Bai~\cite{Bai20092434} & 60& 82 & 37 &43 &1.47  \\
	Zafari~\cite{zafari2015segmentation} &  60 & 73 & \textbf{98} &73& 2.13 \\
	Wen~\cite{curv-th} &  60 & 91   & 64 &60& 1.79  \\
	Dai~\cite{curve-area} &  60 & 63        &98  &62& 2.61  \\
	Samma~\cite{bnd-skeleton} &  60 & 67    & 82 &59& 2.26  \\
	Wang~\cite{skeleton} & 60 & 80  & 88 &72& 1.96  \\
	Kumar~\cite{chord} &  60 & 58   & \textbf{98} &57& 2.71  \\
	Farhan~\cite{farhan2013novel}   &  60 & 82      & 19 &19& 1.91  \\
	\bottomrule
\end{tabular}
\caption{Comparison of the performance of concave points detection methods on the synthetic dataset.}\label{tab:1}
\end{table} 

%------------------------------------------

\begin{figure*}[h]
\begin{center}
	\begin{tabular}{c c c }
		{\includegraphics [width=.23\textwidth]{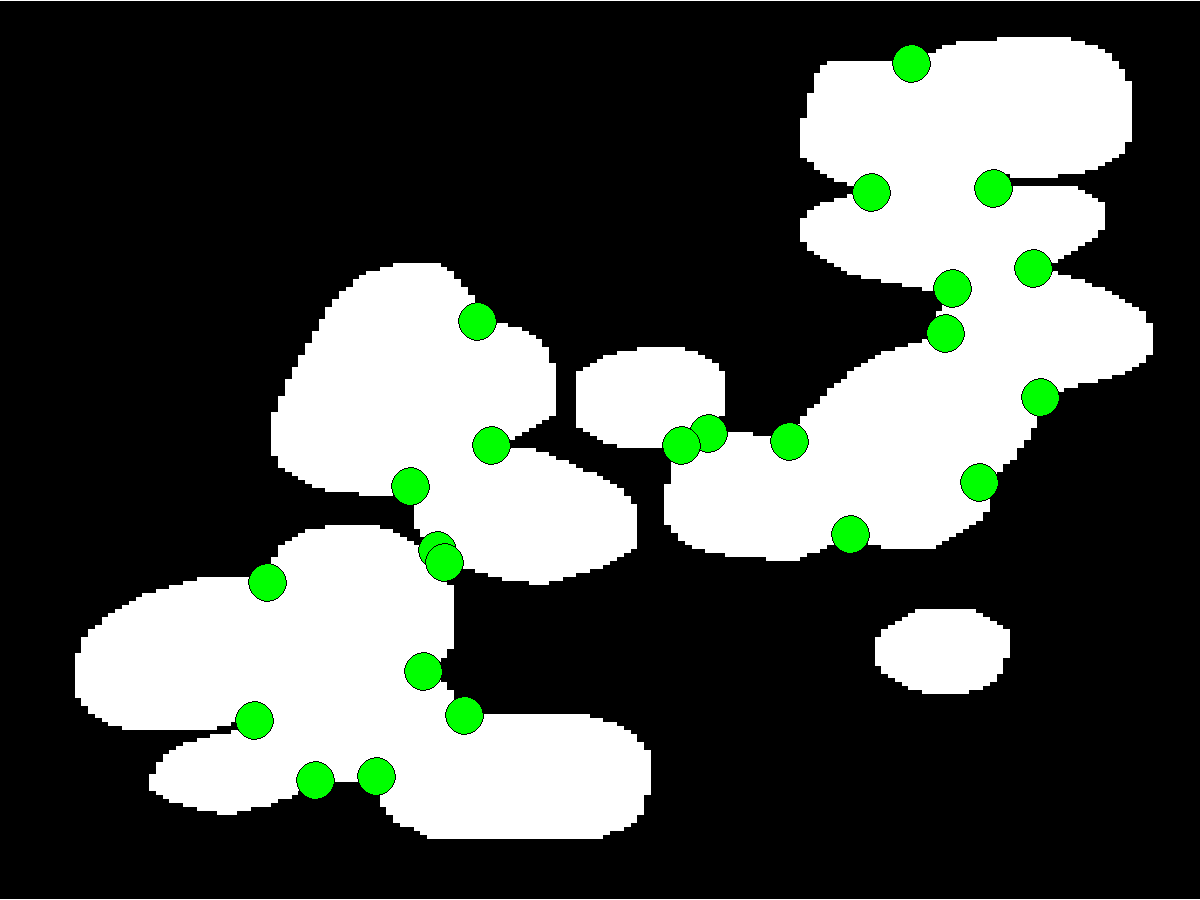}}& 
		{\includegraphics [width=.23\textwidth]{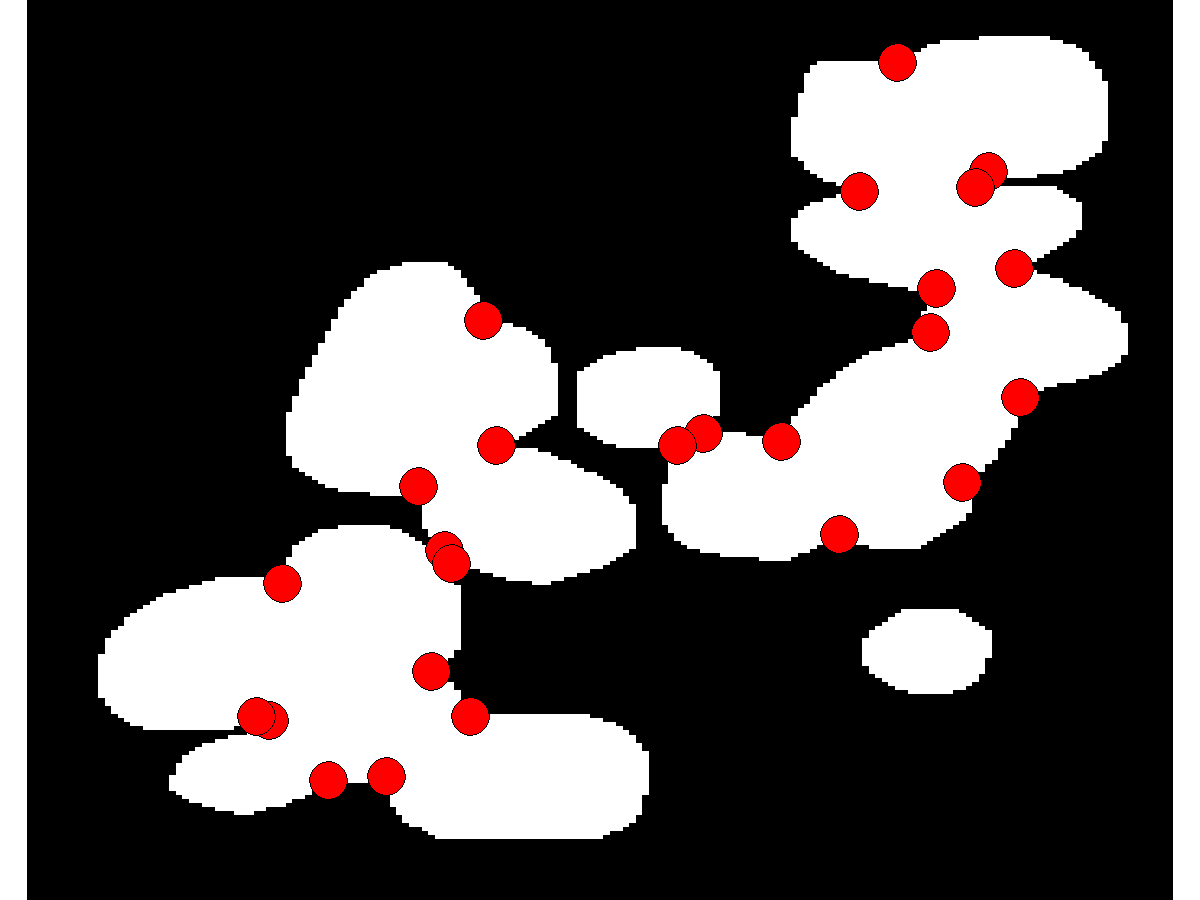}}&
		{\includegraphics [width=.23\textwidth]{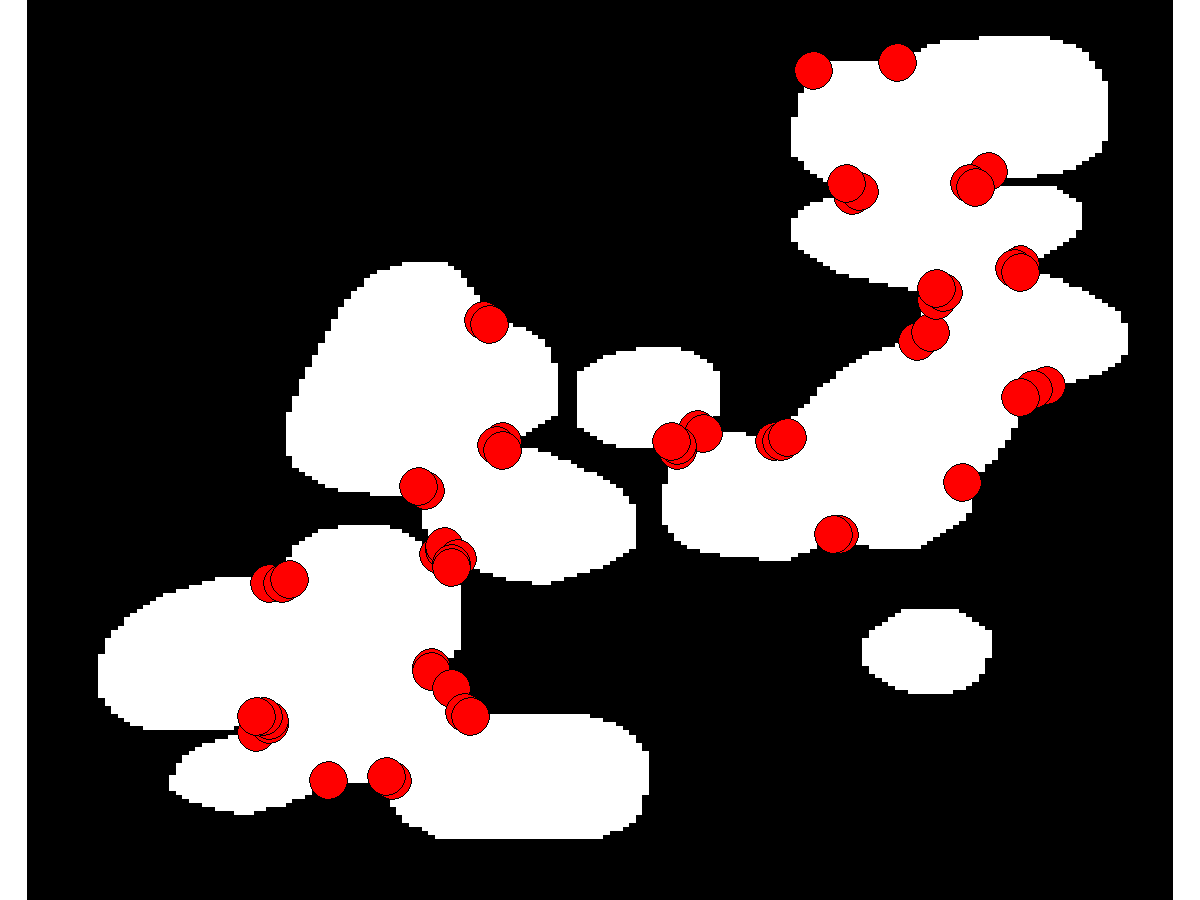} }\\
		(a) & (b) & (c)\\
	\end{tabular}
	
	\begin{tabular}{c c c c }
		{\includegraphics [width=.23\textwidth]{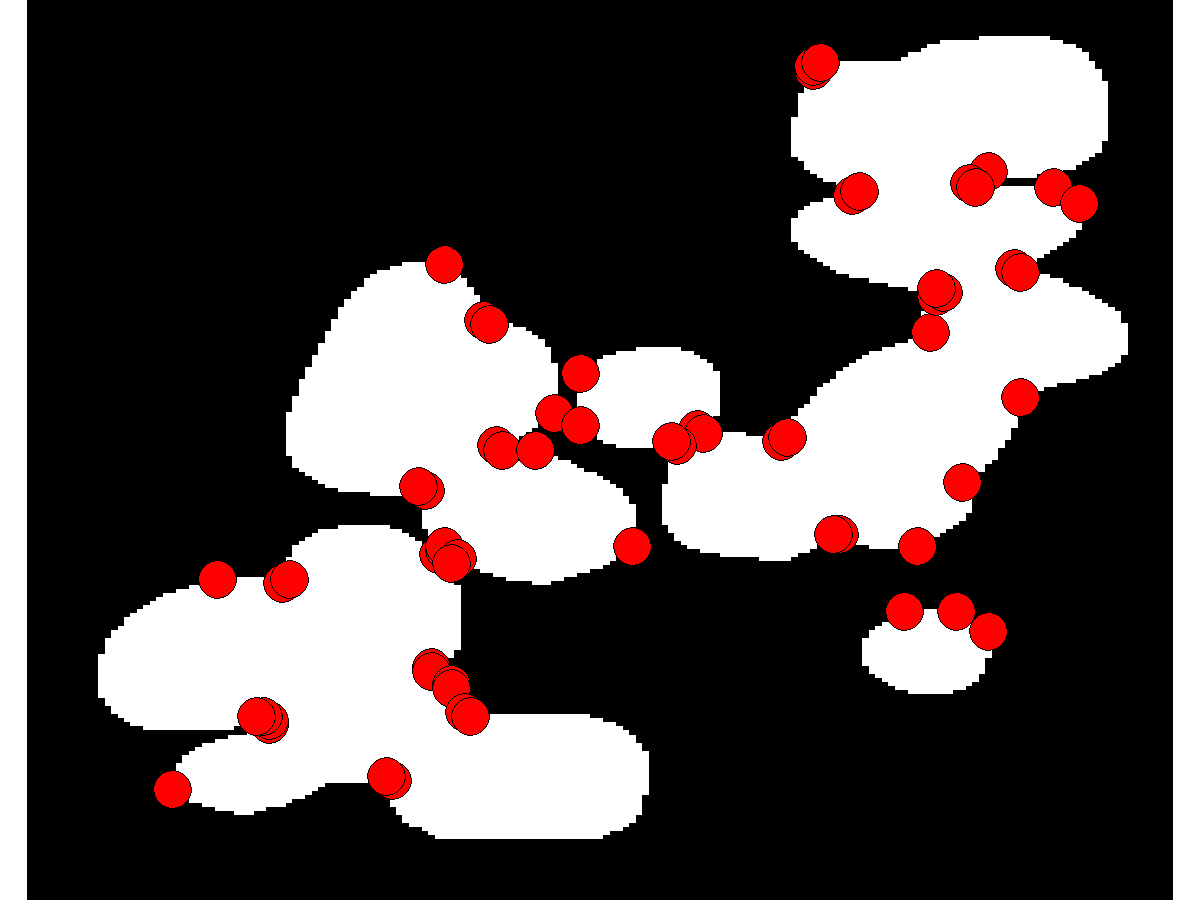} } &
		{\includegraphics [width=.23\textwidth]{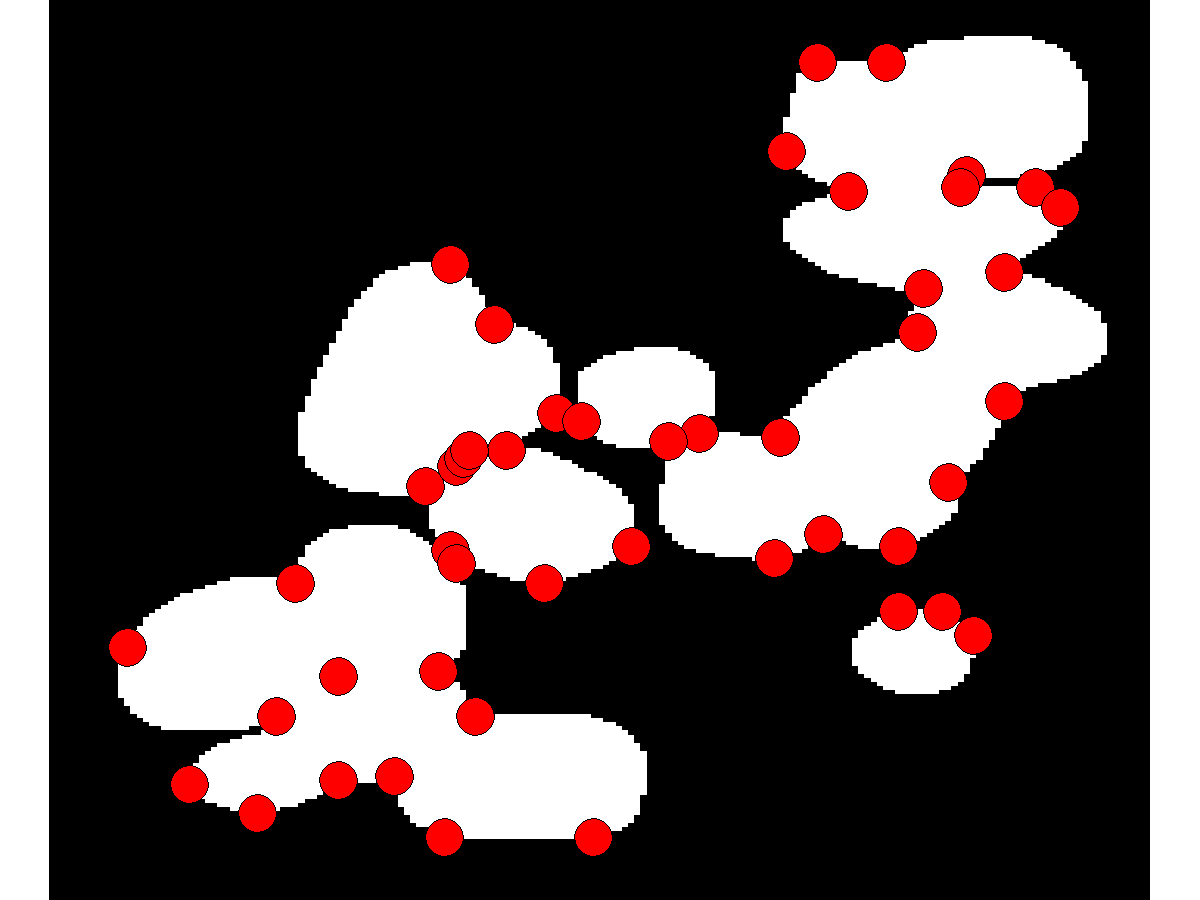}}&
		{\includegraphics [width=.23\textwidth]{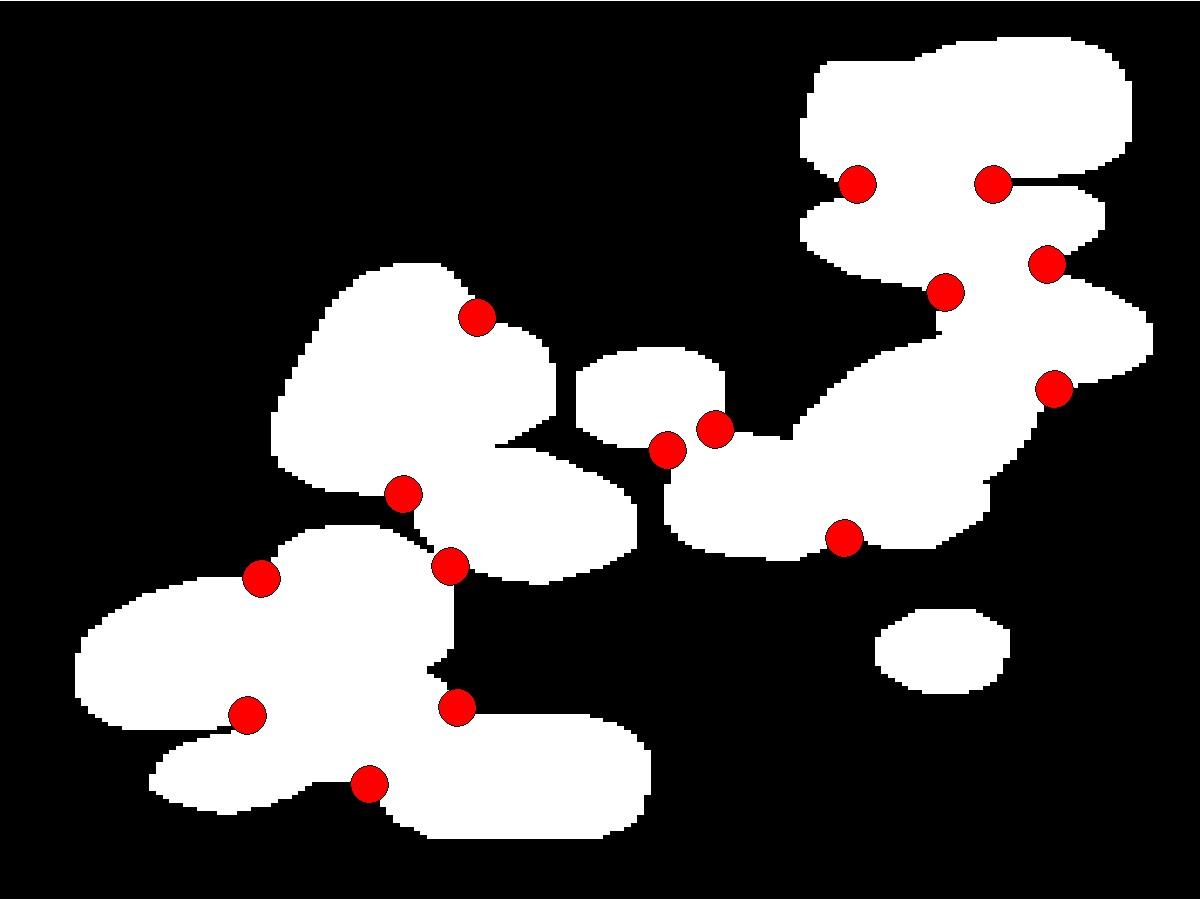}} &
		{\includegraphics [width=.23\textwidth]{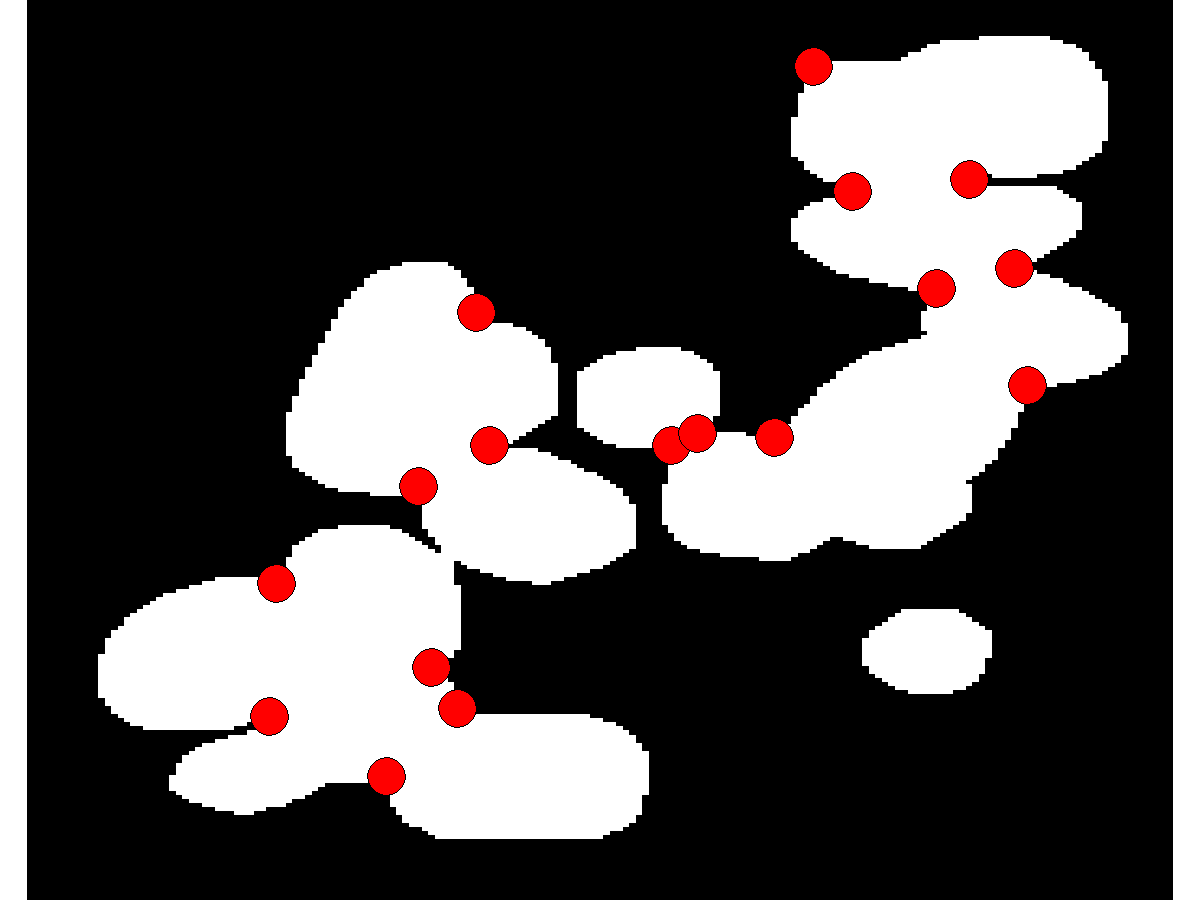}}\\
		(d) & (e) & (f) & (g)\\
	\end{tabular}
	\begin{tabular}{c c c c}
		{\includegraphics [width=.23\textwidth]{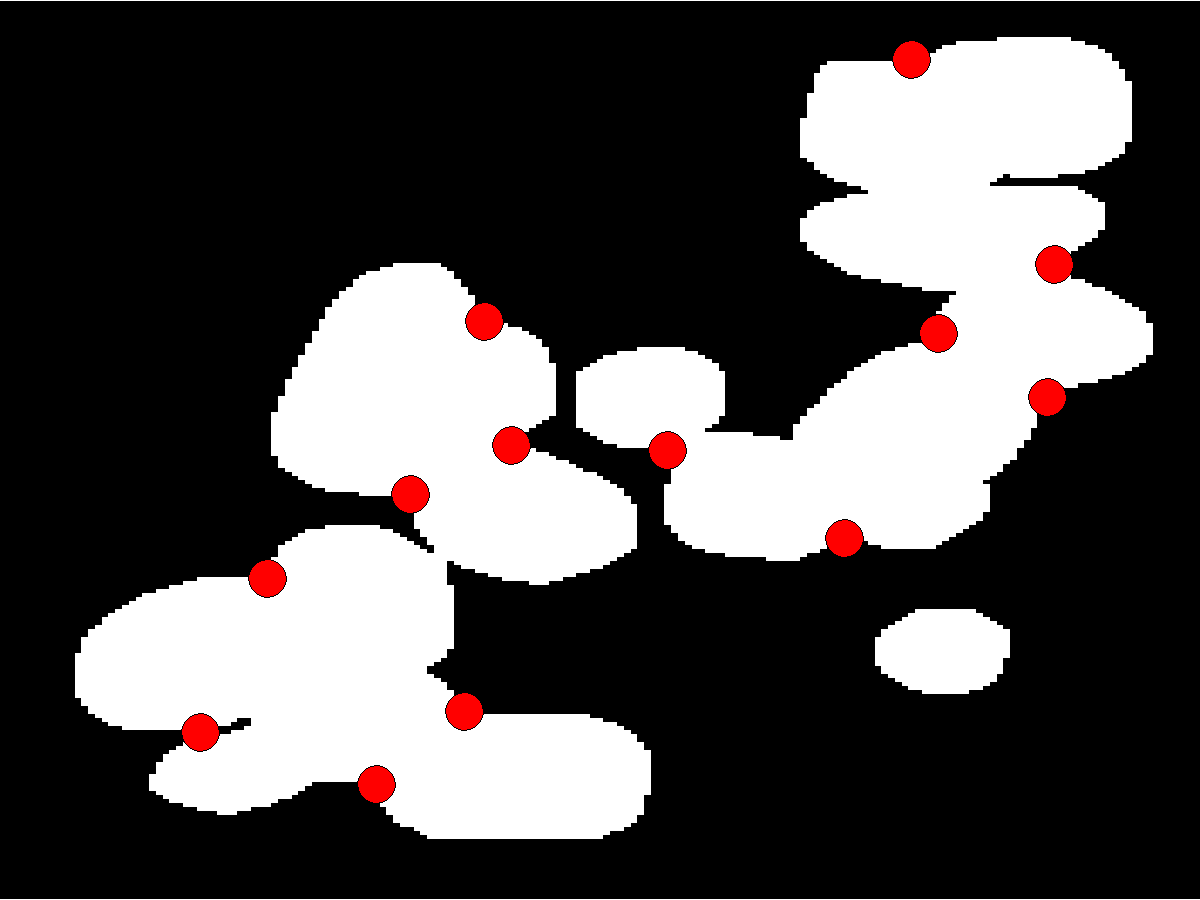}}& 
		{\includegraphics [width=.23\textwidth]{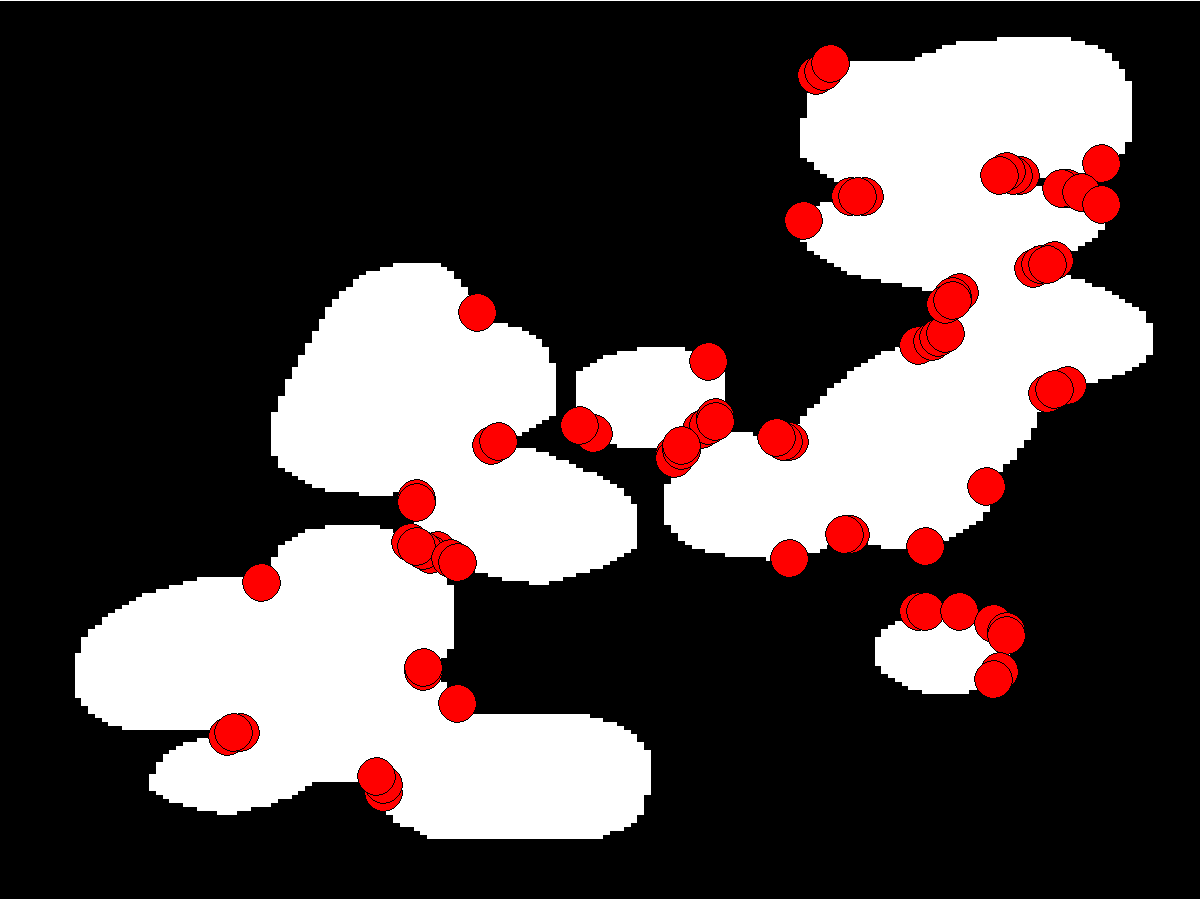}} &
		{\includegraphics [width=.23\textwidth]{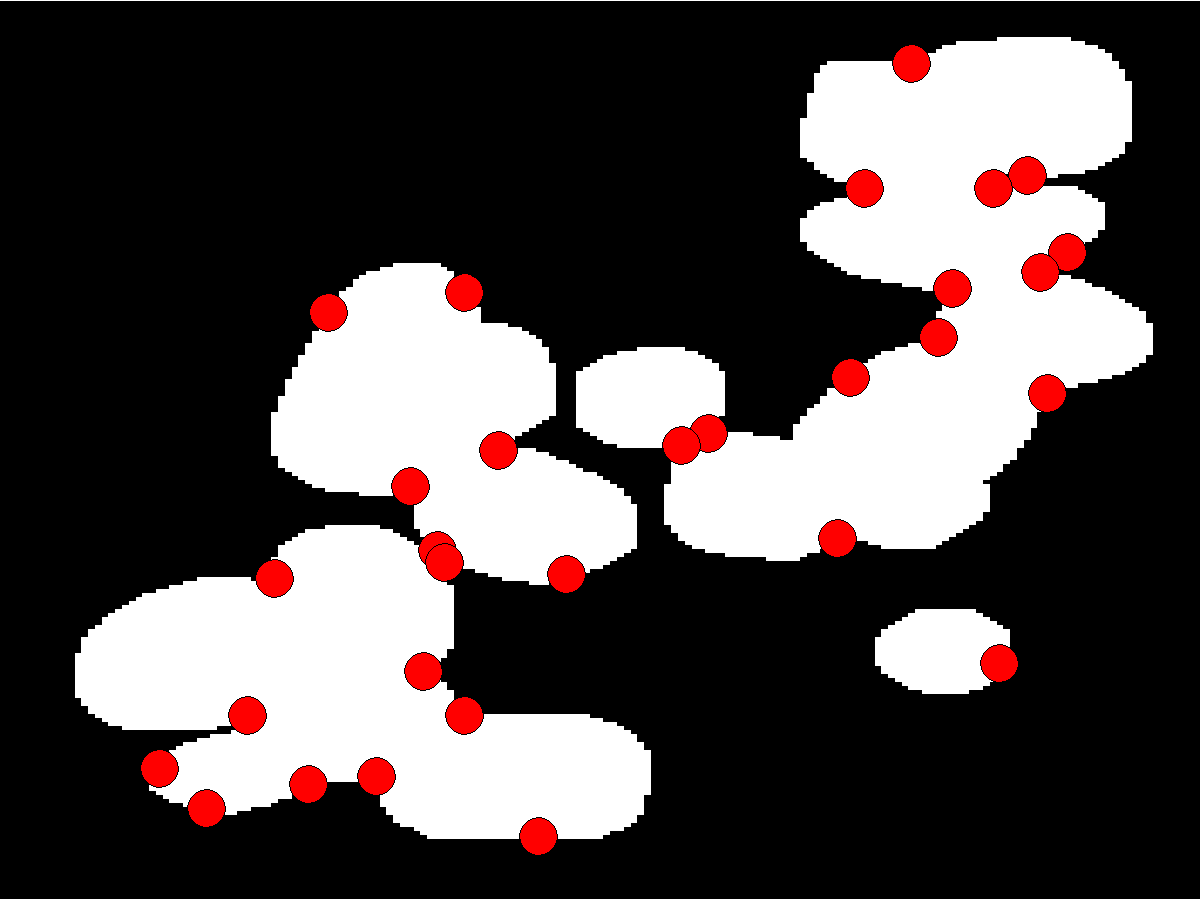}} &
		{\includegraphics [width=.24\textwidth]{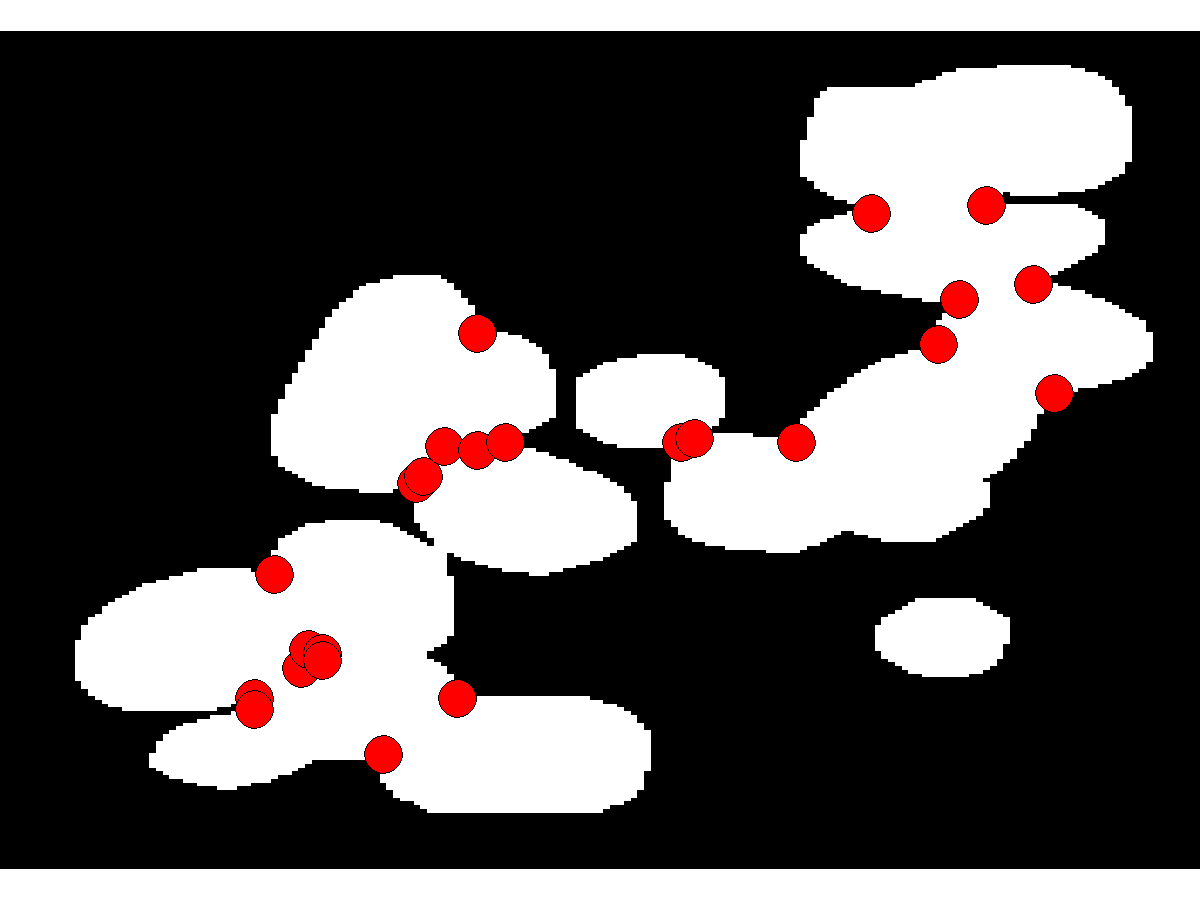}} \\
		(h) & (i) & (j) & (k) \\
	\end{tabular}
	{\caption[]{Comparison of the performance of the concave points detection methods on the nanoparticles dataset: (a) Ground Truth; (b) Prasad+Zhang~\cite{Zafari2017}; (c) Zhang~\cite{bubble}; (d) Bai~\cite{Bai20092434}; (e) Wen~\cite{curv-th}; (f) Zafari~\cite{zafari2015segmentation}; (g) Dai~\cite{curve-area}; (h) Kumar~\cite{concavity}; (i)~Farhan~\cite{farhan2013novel}; (j) Wang~\cite{skeleton}; (k) Samma~\cite{bnd-skeleton}.}\label{fig:4}}
\end{center}
\end{figure*}

Fig.~\ref{fig:4} shows example results of the concave points detection methods applied to a patch of a nanoparticles image. It can be seen that while Zafari~\cite{zafari2015segmentation}, Dai~\cite{curve-area}, Kumar~\cite{concavity} and Samma~\cite{bnd-skeleton} suffer from false negatives and Zhang~\cite{bubble}, Bai~\cite{Bai20092434}, Wen~\cite{curv-th} and Farhan~\cite{farhan2013novel} suffer from false positives, Prasad+Zhang~\cite{Zafari2017} suffers 
from neither false positives nor false negatives.

%\FloatBarrier
%-----------------------------------------Figure 4(example of comparison for CPD)

%\FloatBarrier  
\subsubsection{Contour estimation} 

Tables~\ref{tab:3} and \ref{tab:4}
show the results of contour estimation with the proposed GP-PF using various types of commonly used covariance functions 
on the nanoparticles dataset using both the ground truth contour evidences and the contour evidences obtained using
the proposed contour evidence extraction method.
As it can be seen the Matern family outperforms the other covariance functions 
with both the ground truth and obtained contour evidences.

%-------------------------------------------- Table 3
\begin{table}[ht!]
\centering
\begin{tabular}{p{2.5cm}>
		{\centering\arraybackslash}p{1cm}>
		{\centering\arraybackslash}p{0.9cm}>
		{\centering\arraybackslash}p{0.9cm}>
		{\centering\arraybackslash}p{0.9cm}>
		{\centering\arraybackslash}p{2.5cm}>
		{\centering\arraybackslash}p{0.1cm}}
	%  \toprule
	\toprule
	\multirow{2}{*}{Methods}& TPR & PPV  & ACC & AJSC\\
	&[\%] &[\%]&[\%]&[\%]\\
	\midrule
	Matern 3/2      & \textbf{96} &  {96}& {{92}} &\textbf{{89}}\\
	
	Matern 5/2 & \textbf{96} & \textbf{{97}}        & \textbf{93} &\textbf{89} \\
	
	Squared exponential & 91  &     92& 84 &85\\
	
	Rational quadratic &95  & 96 &92 & 88\\
	\bottomrule
\end{tabular}
\caption[]{The performance of the proposed GP-PF method with different covariance functions on the ground truth contour evidences.}\label{tab:3}
\end{table}

%-------------------------------------------- Table 2
\begin{table}[ht!]
\centering
\begin{tabular}{p{2.5cm}>
		{\centering\arraybackslash}p{1cm}>
		{\centering\arraybackslash}p{0.9cm}>
		{\centering\arraybackslash}p{0.9cm}>
		{\centering\arraybackslash}p{0.9cm}>
		{\centering\arraybackslash}p{2.5cm}>
		{\centering\arraybackslash}p{0.1cm}}
	%  \toprule
	\toprule
	\multirow{2}{*}{Methods}& TPR & PPV  & ACC & AJSC\\
	&[\%] &[\%]&[\%]&[\%]\\
	\midrule
	Matern 3/2      & \textbf{87} & \textbf{83} & \textbf{74} & \textbf{87}\\
	
	Matern 5/2 & \textbf{87 }& \textbf{83}  &\textbf{74}  & \textbf{87}\\
	
	Squared exponential & 83  &     78      & 68 & 83\\
	
	Rational quadratic & 87 & 82  & 73 & 85 \\
	\bottomrule
\end{tabular}
\caption[]{The performance of the proposed GP-PF method with different covariance functions on the obtained contour evidences.}\label{tab:4}
\end{table}

%-------------------------------------------- Table 4
\begin{table}[ht!]
\centering
\begin{tabular}{p{1.8cm}>
		{\centering\arraybackslash}p{1.5cm}>
		{\centering\arraybackslash}p{0.9cm}>
		{\centering\arraybackslash}p{0.9cm}>
		{\centering\arraybackslash}p{0.9cm}>
		{\centering\arraybackslash}p{1.5cm}>
		{\centering\arraybackslash}p{0.1cm}}
	%  \toprule
	\toprule
	\multirow{2}{*}{Methods}& TPR & PPV  & ACC & AJSC\\
	&[\%] &[\%]&[\%]&[\%]\\
	\midrule
	GP-PF   & \textbf{96} & \textbf{96}     & \textbf{92} &\textbf{90}\\
	
	%       GP-IPF & 88 &   88      & 80 &88\\
	
	BS & 92 &     92      & 86 &89\\
	
	LESF & 94& 95 & 89& 88\\
	\bottomrule
\end{tabular}
\caption[]{The performance of the contour estimation methods on the ground truth contour evidences.}\label{tab:5}
\end{table}

To demonstrate the effect of the more accurate contour estimation on the segmentation of partially overlapping objects, 
the performance of the GP-PF method was compared to the BS and the LESF methods on the nanoparticle dataset.
Table~\ref{tab:5} shows the segmentation results produced by the contour estimation methods applied to the ground truth contour evidences and Table~\ref{tab:6} presents their performance with the obtained contour evidences.

%--------------------------------------------Table 6
\begin{table}[h!]
\centering
\begin{tabular}{p{1.5cm}>
		{\centering\arraybackslash}p{1.5cm}>
		{\centering\arraybackslash}p{0.9cm}>
		{\centering\arraybackslash}p{0.9cm}>
		{\centering\arraybackslash}p{0.9cm}>
		{\centering\arraybackslash}p{2.5cm}>
		{\centering\arraybackslash}p{0.1cm}}
	%  \toprule
	\toprule
	\multirow{2}{*}{Methods}& TPR & PPV  & ACC & AJSC\\
	&[\%] &[\%]&[\%]&[\%]\\
	\midrule
	GP-PF   &\textbf{87} &  \textbf{{83}}   & \textbf{74}&\textbf{87}\\ 
	%       GP-IPF & 78 &   72      & 60& 86 \\
	
	BS & 82 &     76      & 66 & 87\\
	
	LESF & 85 & {83} & {71}& 87 \\
	\bottomrule
\end{tabular}
\caption{The performance of the contour estimation methods on the obtained contour evidences.}\label{tab:6}
\end{table}

The results show the advantage of the proposed GP-PF method over the other methods when applied to the ground truth contour evidences. In all terms the proposed contour estimation method achieved the highest performance.
The results on the obtained contour evidences (Table~\ref{tab:6}) confirm the out-performance of the GP-PF method over the others with the highest values. 

In general, the LESF contour estimation takes advantage of
objects with an elliptical shape which can lead to desirable contour estimation if the objects are in the form of ellipses. However, in situations where objects are not close to elliptical shapes, the method fails to estimate the exact contour of the objects. The BS contour estimation is not limited to specific object shapes, but requires to learn many contour functions with prior shape information. The proposed Gaussian process contour estimation can resolve the actual contour of the objects without any strict assumption on the object's shape.

\begin{figure}[h]
\begin{center}
	\begin{tabular}{c c c c }                   
		\frame{\includegraphics [width=0.45\linewidth]{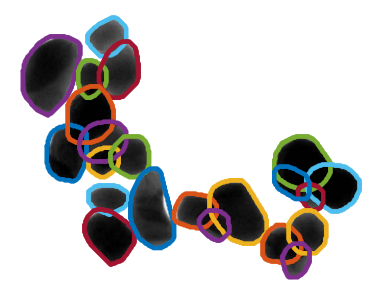}
		}&
		\frame{\includegraphics [width=0.45\linewidth]{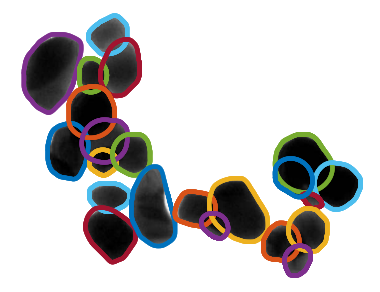}
		}\\
		(a) & (b)\\
		\frame{\includegraphics [width=0.45\linewidth]{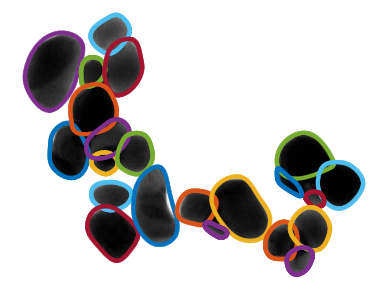}
		}&
		\frame{\includegraphics [width=0.45\linewidth]{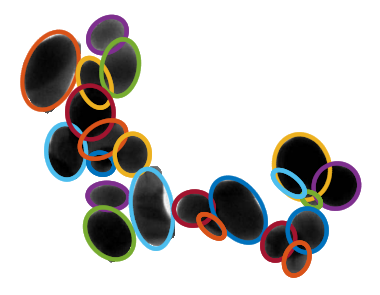}
		}\\
		(c) & (d) \\         
	\end{tabular}
	\caption{Comparison of the performance of contour estimation methods on the ground truth contour evidences: (a) Ground truth full contour; (b) GP-PF; (c) BS; (d) LESF.} \label{fig:5}
\end{center}
\end{figure}

%------------------------------------
Fig.~\ref{fig:5} shows example results of contour estimation methods applied to a slice of nanoparticles dataset. 

It can be seen that BS tends to underestimate the shape of the objects. With small contour evidences, the BS method results in a small shape  while LESF can result in larger shapes. The proposed GP-PF method estimates the objects boundaries more precisely.

\begin{figure*}[htp]
\begin{tabular}{c c c }       
	{\includegraphics [width=0.3\linewidth]{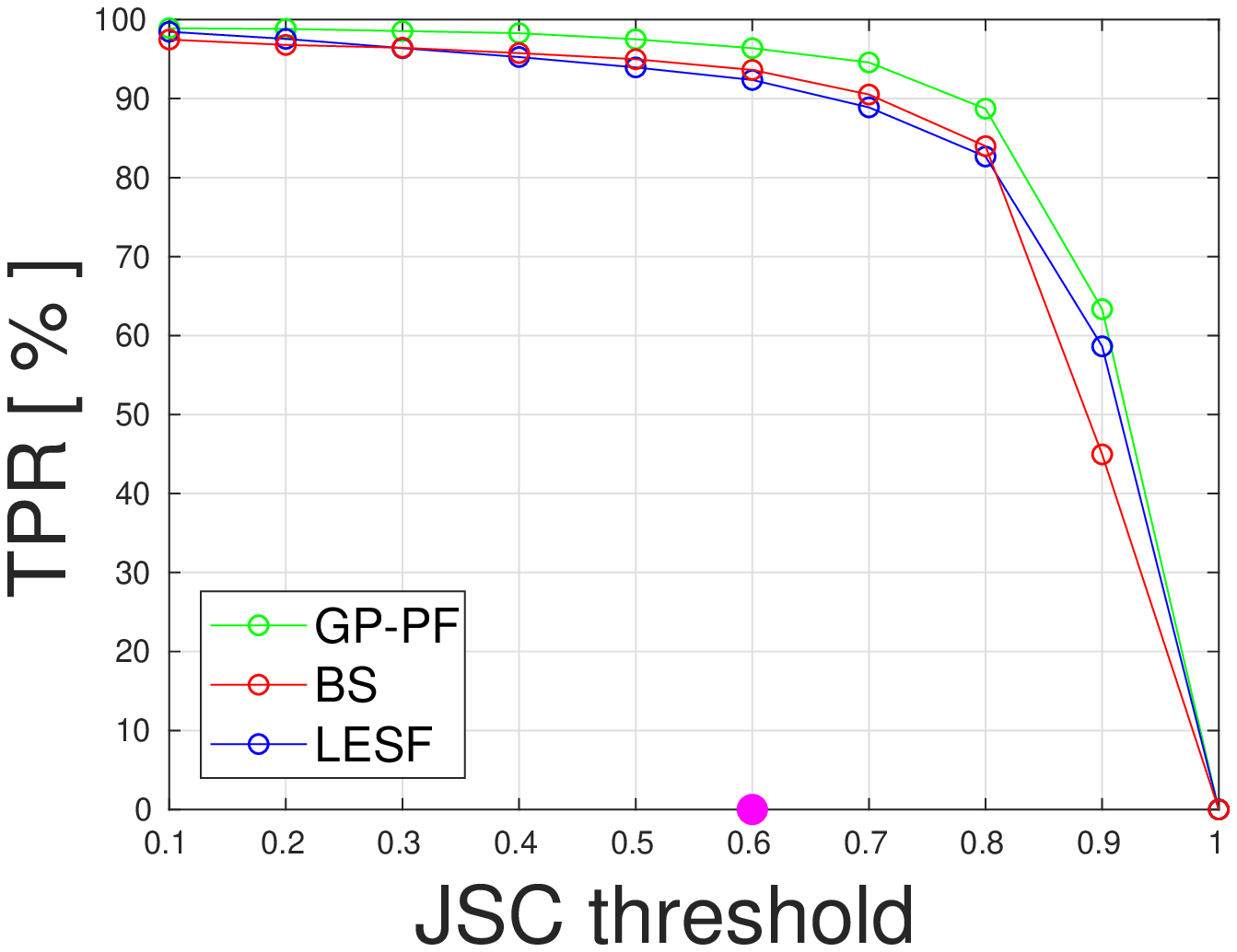}
	}&
	{
		\includegraphics [width=0.3\linewidth]{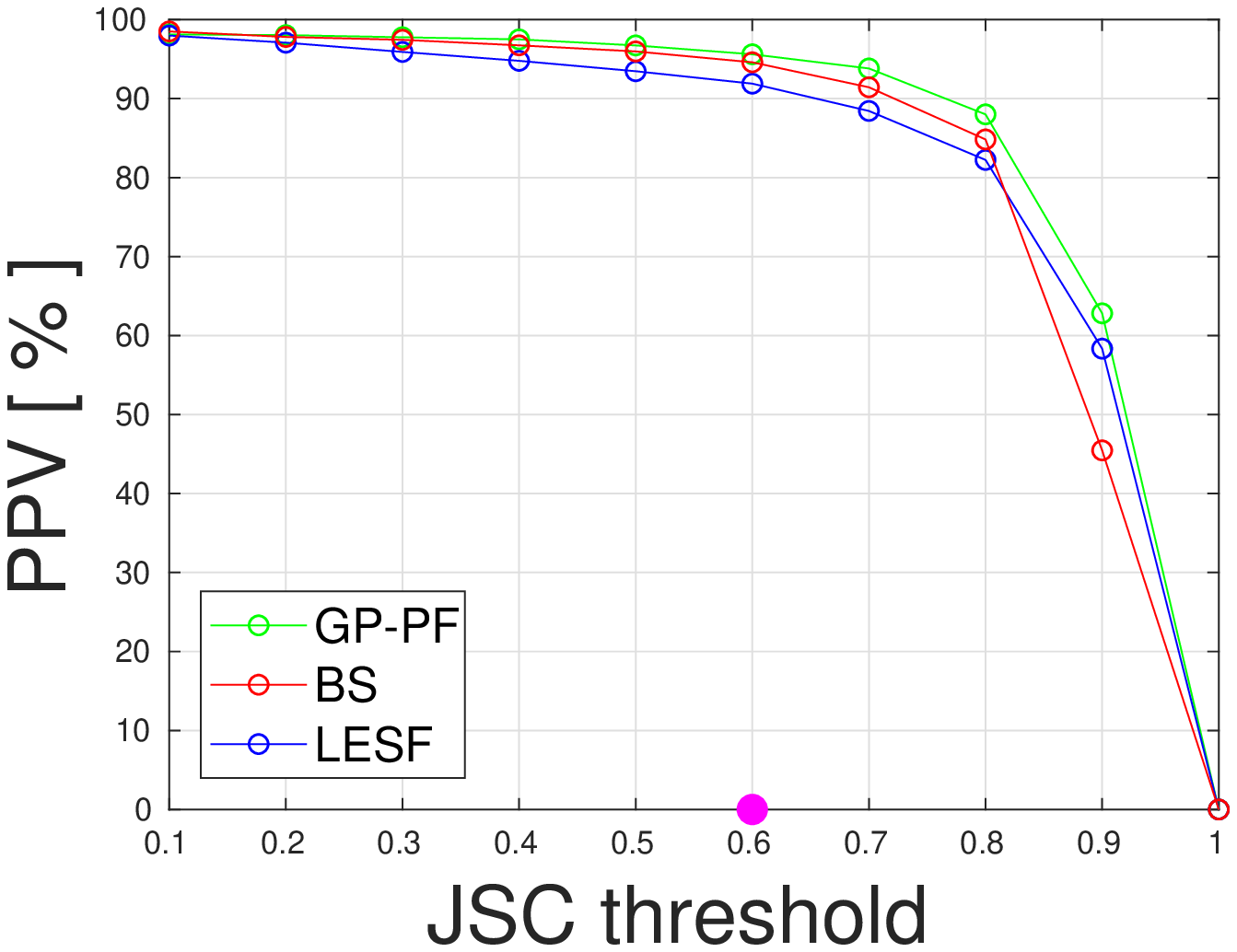}
	}&
	{
		\includegraphics [width=0.3\linewidth]{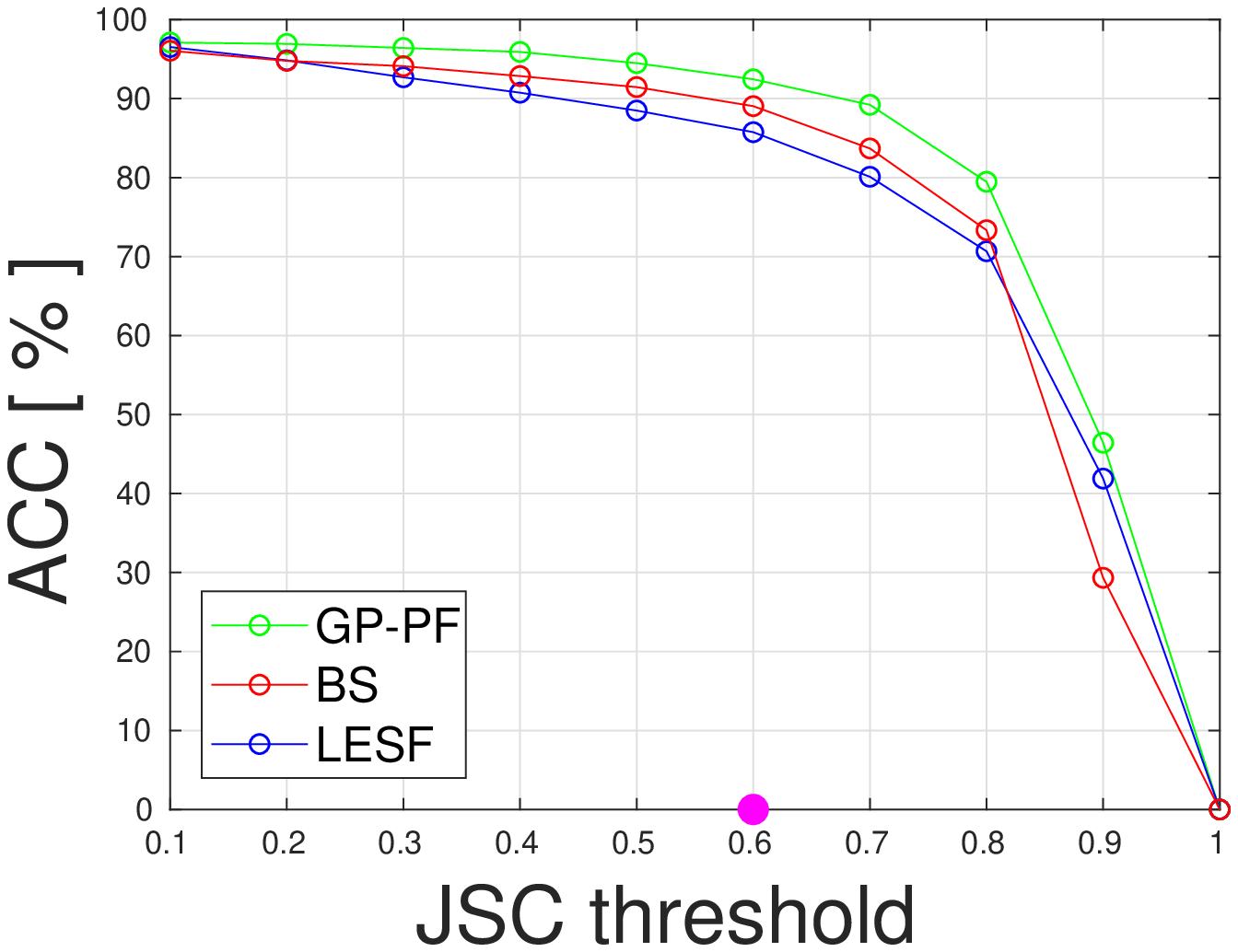}
	}\\
	(a) & (b) & (c)\\
\end{tabular}
\caption[moving]{Evaluation results for the contour estimation methods with the different JSC thresholds value applied to the ground truth contour evidences: (a) TPR; (b) PPV; (c) ACC.}
\label{fig:6}
\end{figure*}

\begin{figure*}[htp]
\begin{tabular}{c c c }       
	{\includegraphics [width=0.3\linewidth]{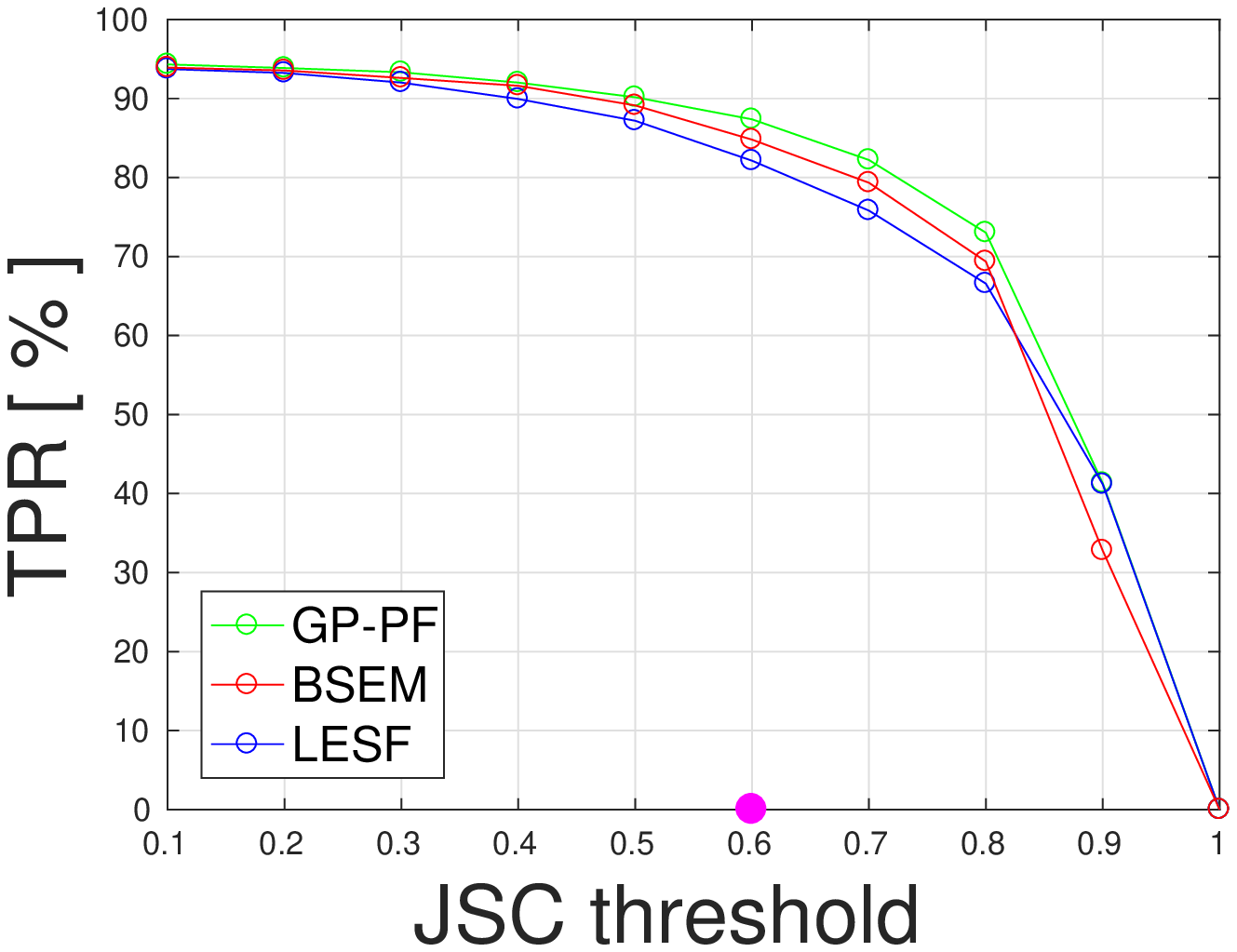}
	}&
	{
		\includegraphics [width=0.3\linewidth]{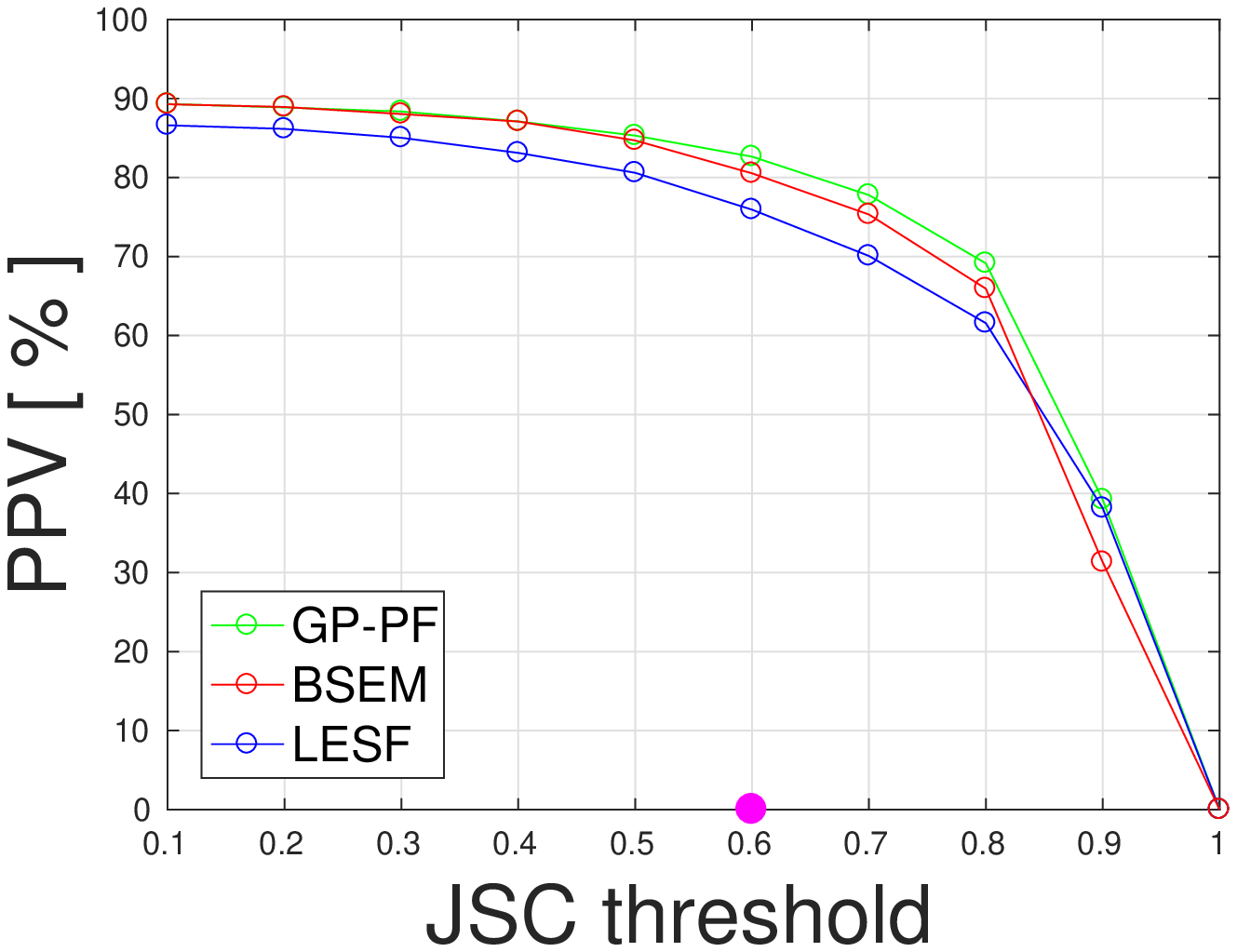}
	}&
	{
		\includegraphics [width=0.3\linewidth]{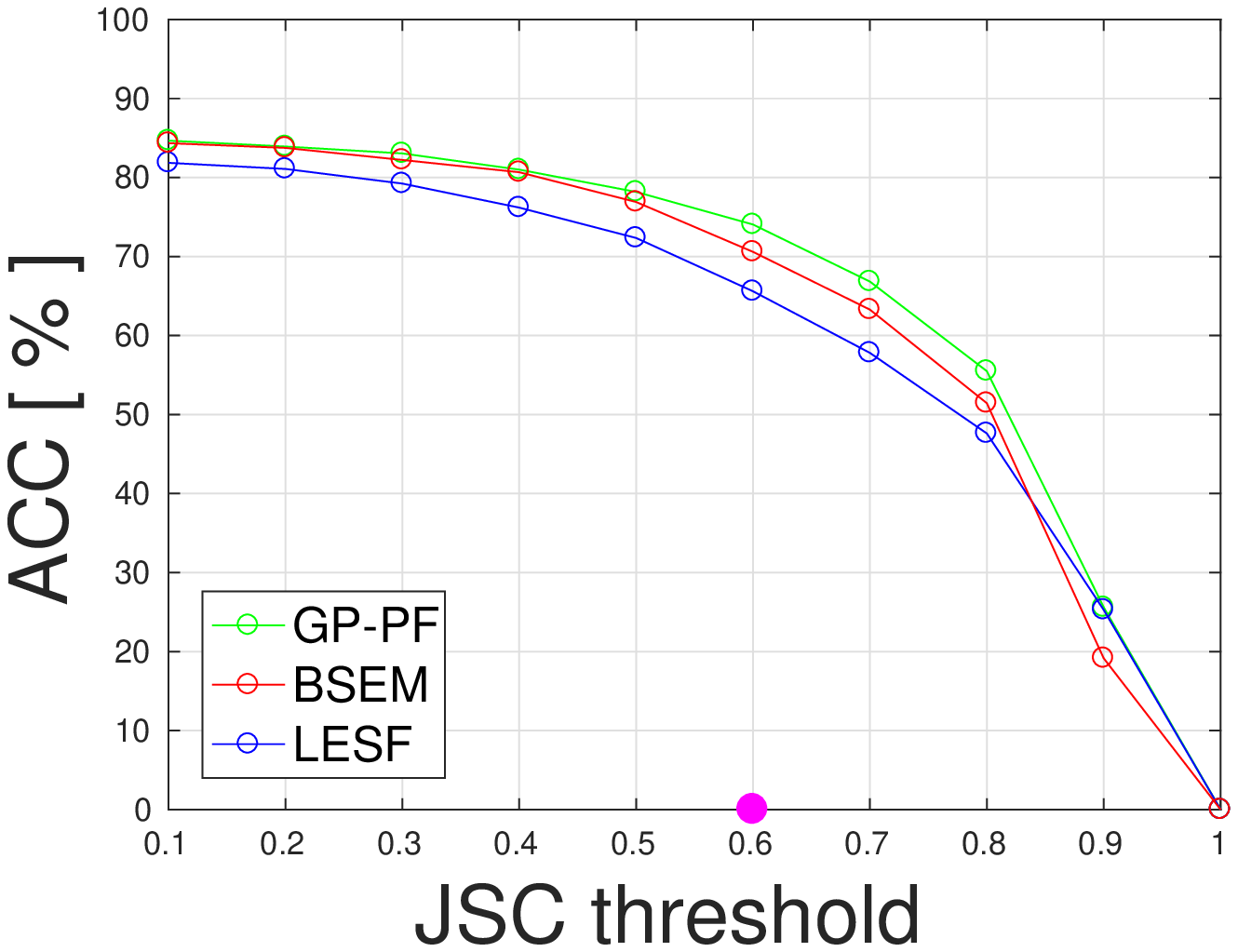}
	}\\
	(a) & (b) & (c)\\
\end{tabular}
\caption[moving]{Evaluation results for the contour estimation methods with the different JSC threshold values applied to the obtained contour evidences: (a) TPR; (b) PPV; (c) ACC.}
\label{fig:7}
\end{figure*}

Figs.~\ref{fig:6} and  \ref{fig:7} show the effect of the Jaccard similarity threshold on the TPR, PPV, and ACC scores with the proposed and competing contour estimation methods. 
As expected, the segmentation performance of all methods degrades when the JSC threshold is increased. However, the threshold value has only the minor effect on the ranking order of the methods and the proposed contour estimation method outperforms the other methods with higher JSC threshold values.

%-------------------------------------Figure 6(comparison of contour estimation threshold)

%-------------------------------------Figure 7(comparison of contour estimation threshold)

\subsubsection{Segmentation of overlapping convex objects}

\begin{figure*}[htp]
\begin{center}
	\begin{tabular}{c c }
		\frame{\includegraphics [width=.45\textwidth]{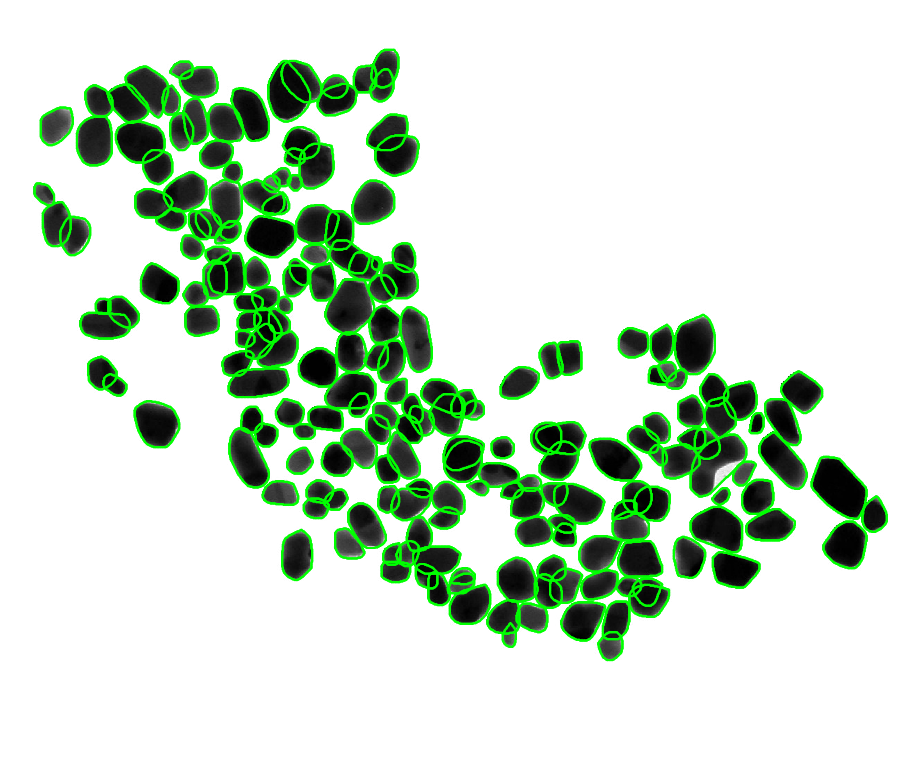}}& 
		\frame{\includegraphics [width=.45\textwidth]{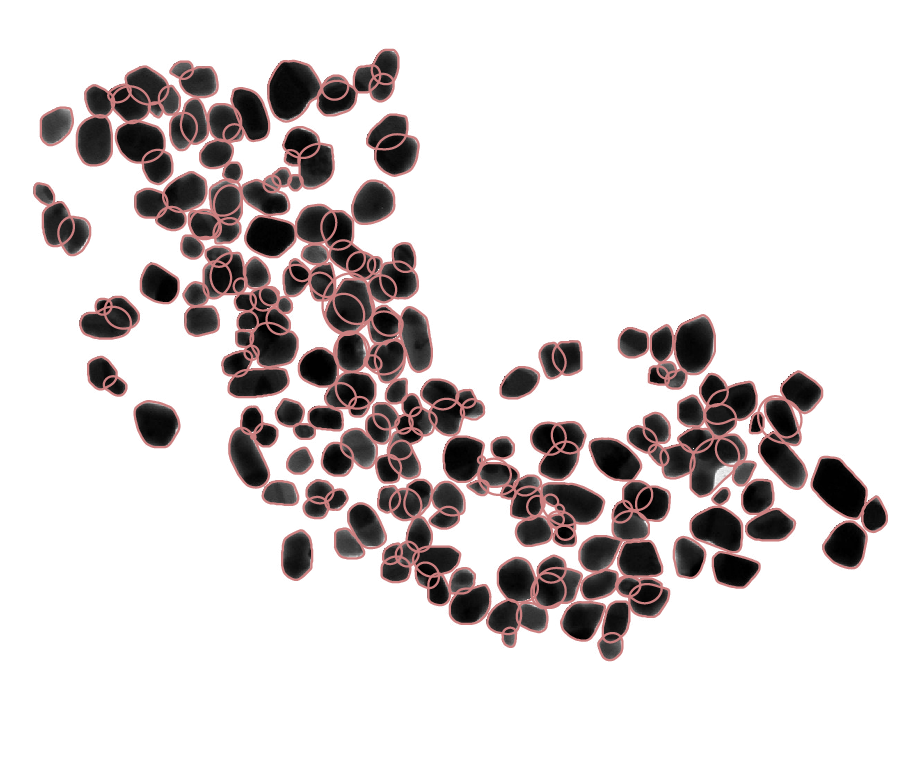}} \\
		(a) & (b)
	\end{tabular}
	\vspace{.1cm}
	\caption[]{An example of the proposed method segmentation result on the nanoparticles dataset: (a) Ground truth; (b) Proposed method.}\label{fig:8}
\end{center}
\end{figure*}

\sloppy The proposed method is organized in a sequential manner where the performance of each step directly impact the performance of the next step. In each step, all the potential methods are compared and the best performer is selected. The performance of the full proposed segmentation method using the Prasad+Zhang concave points detection and the proposed GP-PF contour estimation methods was compared to three existing methods:  seed point based contour evidence extraction and contour estimation (SCC)~\cite{7300433},  nanoparticles segmentation (NPA)~\cite{npa}, and concave-point extraction and contour segmentation (CECS)~\cite{bubble}.
These methods are particularly chosen as previously applied for segmentation of overlapping convex objects. 
Fig.~\ref{fig:8} shows an example of segmentation result for the proposed method. 

%-----------------------------------------------Figure 8
%--------------------------------------------------------
%------------------------------------------- Figure 10
\begin{figure*}[ht]
\begin{tabular}{c c c }       
	{\includegraphics [width=0.3\linewidth]{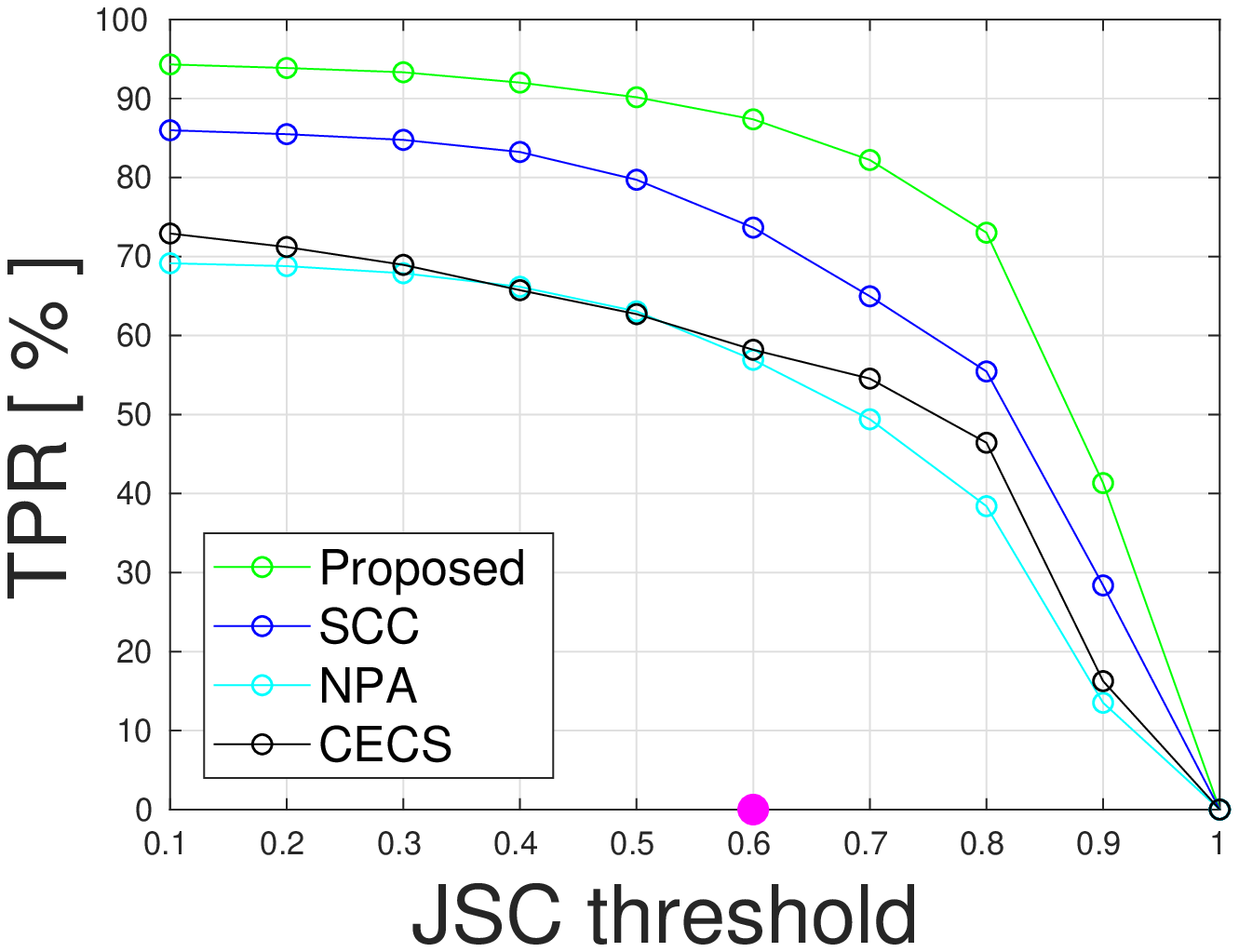}
	}&
	{
		\includegraphics [width=0.3\linewidth]{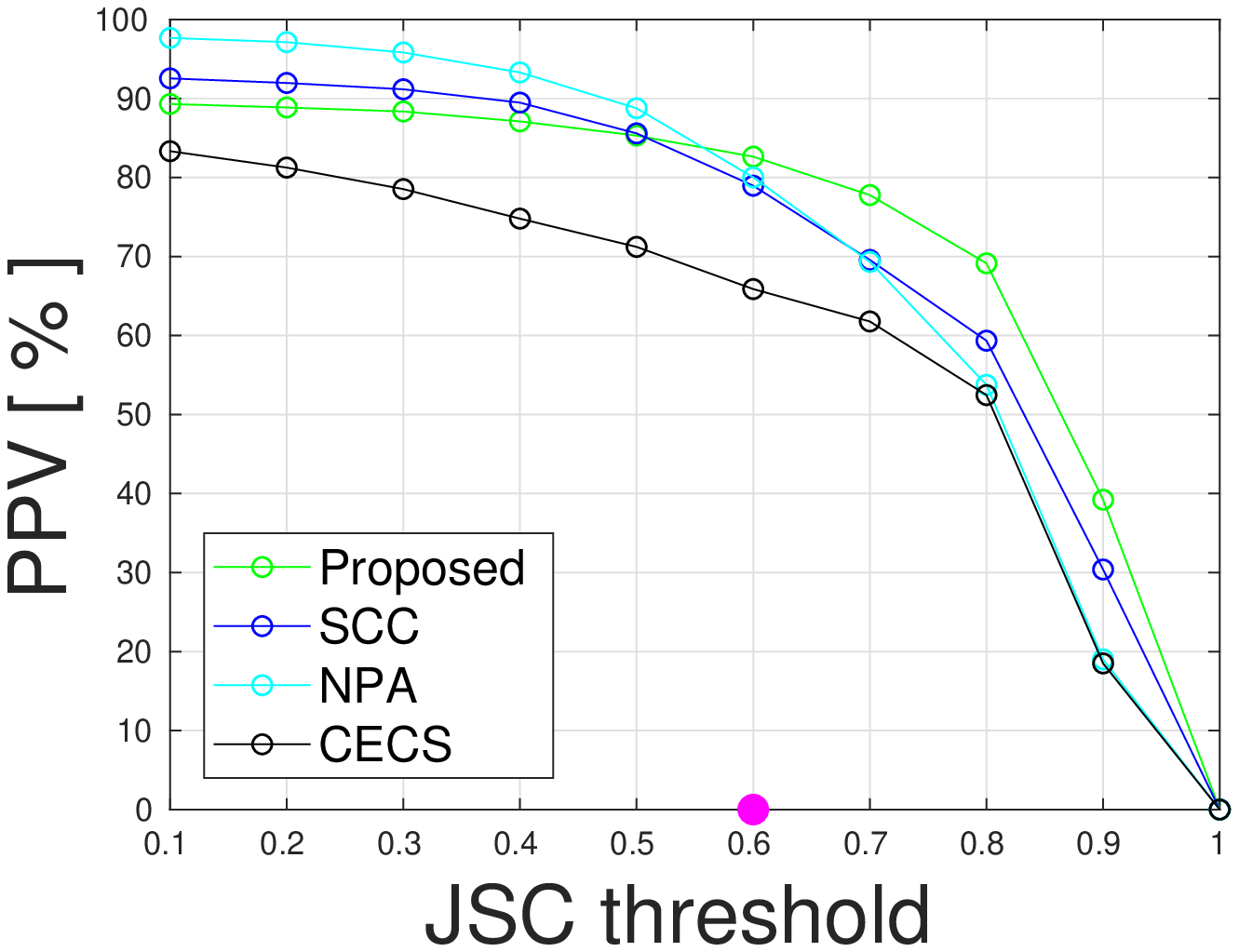}
	}&
	{
		\includegraphics [width=0.3\linewidth]{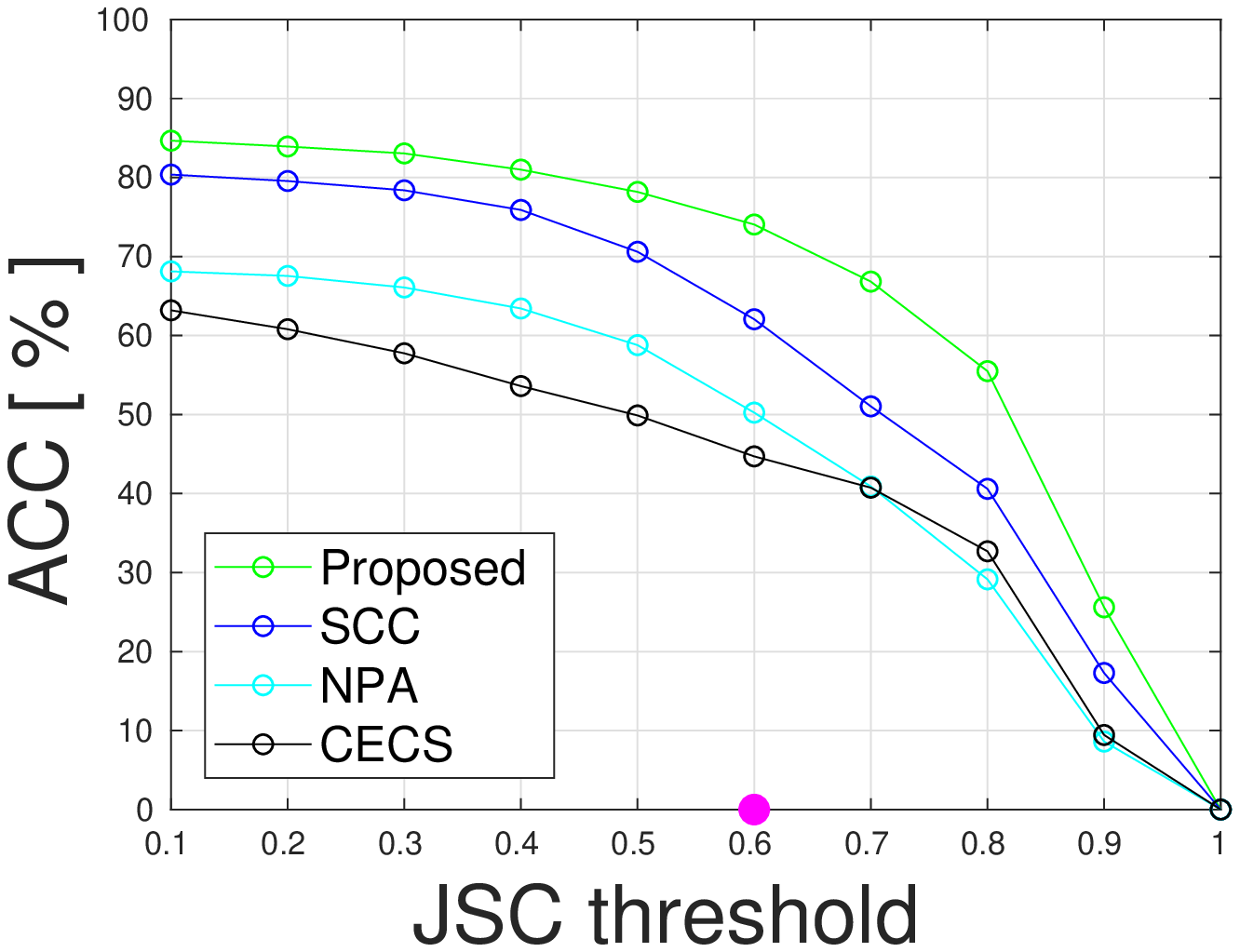}
	}\\
	(a) & (b) & (c)\\
\end{tabular}
\caption[moving]{Evaluation results for the segmentation methods with the different JSC threshold values: (a) TPR; (b) PPV; (c) ACC.}
\label{fig:09}
\end{figure*}

%-----------------------------------------

The corresponding performance statistics of the competing methods applied to the nanoparticles dataset are shown in Table~\ref{tab:7}. 
As it can be seen, the proposed method outperformed the other four methods with respect to TPR, PPV, ACC, and AJSC. 
The highest TPR, PPV, and ACC indicates its superiority with respect to the resolved overlap ratio.

Fig \ref{fig:09} shows the effect of the Jaccard similarity threshold on the TPR, PPV, and ACC scores with the proposed segmentation and competing methods.
As it can be seen, the performance of all segmentation methods degrades when the JSC threshold is increased, 
and the proposed segmentation method outperforms the other methods especially with higher JSC threshold values.

%------------------------------------------- Table 7
\begin{table}[ht!]
	\centering
	\begin{tabular}{p{2cm}>{\centering\arraybackslash}p{0.9cm}>
			{\centering\arraybackslash}p{0.7cm}>
			{\centering\arraybackslash}p{0.7cm}>
			{\centering\arraybackslash}p{0.7cm}>
			{\centering\arraybackslash}p{0.7cm}>
			{\centering\arraybackslash}p{0.3cm}}
		%  \toprule
		\toprule
		\multirow{2}{*}{Methods}&  TPR & PPV  & ACC & AJSC \\
		&[\%]&[\%]&[\%]&[\%]\\
		\midrule
		{Proposed }  & \textbf{87 } &   \textbf{83}     & \textbf{74}& \textbf{87} & \\
		SCC              & 74 & 79 & 62 & 82&\\%
		NPA              & 57 & {80} & 50& 81& \\
		CECS          & 58 & 66 & 45& 83&\\
		
		\bottomrule
	\end{tabular}
	\caption{Comparison of the performance of the segmentation methods on the nanoparticles dataset.}\label{tab:7}
\end{table}
%-------------------------------------------

%\subsection{Computation time}
%
%\noindent
%The proposed method was implemented in MATLAB, using a PC with a 3.20 GHz CPU and 8 GB of RAM. 
%The source code will be available at the project web page~\cite{comphiweb}. 
%With the selected combination of parameters, the computational time was 200 seconds per image of size 1100 $\times$ 1200
%pixels while SCC demanded 77 seconds, SCC 150 seconds, and NPA 200 seconds. 
%The computational time breakdown was as follows: concave points detection 2\%, segment grouping 85\%, and contour estimation 13\%. However, it should be noted that the method performance was not optimized and the computation time could be improved.

\section{Conclusions}\label{sec:conclusion}

\noindent
This paper presented a method to resolve overlapping convex objects in order
to segment the objects in silhouette images.
The proposed method consisted of three main steps: pre-processing to produce binarized images, contour evidence extraction to detect the visible part of each object, and contour estimation to estimate the final objects contours. 
The contour evidence extraction is performed by detecting concave points and solving the contour segment grouping using the branch and bound algorithm. 
Finally, the contour estimation is performed using Gaussian process regression. 
To find the best method for concave points detection, comprehensive evaluation of different approaches was made.
The experiments showed that the proposed method achieved high detection and segmentation accuracy and outperformed four competing methods on the dataset of nanoparticles images.
The proposed method relies only on edge information and can be applied also to other segmentation problems 
where the objects are partially overlapping and have an approximately convex shape.

%-----------------------------------------------------------------------------------------------------
\FloatBarrier
\bibliographystyle{spmpsci} 
\bibliography{references}

\begin{thebibliography}{10}
\providecommand{\url}[1]{{#1}}
\providecommand{\urlprefix}{URL }
\expandafter\ifx\csname urlstyle\endcsname\relax
  \providecommand{\doi}[1]{DOI~\discretionary{}{}{}#1}\else
  \providecommand{\doi}{DOI~\discretionary{}{}{}\begingroup
  \urlstyle{rm}\Url}\fi

\bibitem{alkohfi}
Al-Kofahi, Y., Lassoued, W., Lee, W., Roysam, B.: Improved automatic detection
  and segmentation of cell nuclei in histopathology images.
\newblock IEEE Transactions on Biomedical Engineering \textbf{57}(4), 841--852
  (2010)

\bibitem{acos_2}
Ali, S., Madabhushi, A.: An integrated region-, boundary-, shape-based active
  contour for multiple object overlap resolution in histological imagery.
\newblock IEEE Transactions on Medical Imaging \textbf{31}(7), 1448--1460
  (2012)

\bibitem{angulo2007stochastic}
Angulo, J., Jeulin, D.: Stochastic watershed segmentation.
\newblock In: Proceedings of the 8th International Symposium on Mathematical
  Morphology, pp. 265--276 (2007)

\bibitem{Arteta}
Arteta, C., Lempitsky, V., Noble, J.A., Zisserman, A.: Learning to detect
  partially overlapping instances.
\newblock In: Proceedings of the IEEE Conference on Computer Vision and Pattern
  Recognition (CVPR), pp. 3230--3237 (2013)

\bibitem{Bai20092434}
Bai, X., Sun, C., Zhou, F.: Splitting touching cells based on concave points
  and ellipse fitting.
\newblock Pattern Recognition \textbf{42}(11), 2434 -- 2446 (2009)

\bibitem{bernardis2010finding}
Bernardis, E., Stella, X.Y.: Finding dots: Segmentation as popping out regions
  from boundaries.
\newblock In: Proceedings of the IEEE Conference on the Computer Vision and
  Pattern Recognition (CVPR), pp. 199--206 (2010)

\bibitem{canny}
Canny, J.: A computational approach to edge detection.
\newblock IEEE Transactions on Pattern Analysis and Machine Intelligence
  \textbf{8}(6), 679--698 (1986)

\bibitem{chang2013invariant}
Chang, H., Han, J., Borowsky, A., Loss, L., Gray, J.W., Spellman, P.T., Parvin,
  B.: Invariant delineation of nuclear architecture in glioblastoma multiforme
  for clinical and molecular association.
\newblock IEEE Transactions on Medical Imaging \textbf{32}(4), 670--682 (2013)

\bibitem{chen2013flexible}
Chen, C., Wang, W., Ozolek, J.A., Rohde, G.K.: A flexible and robust approach
  for segmenting cell nuclei from 2d microscopy images using supervised
  learning and template matching.
\newblock Cytometry Part A \textbf{83}(5), 495--507 (2013)

\bibitem{4671118}
Cheng, J., Rajapakse, J.: Segmentation of clustered nuclei with shape markers
  and marking function.
\newblock IEEE Transactions on Biomedical Engineering, \textbf{56}(3), 741--748
  (2009)

\bibitem{Choi10asurvey}
Choi, S.S., Cha, S.H., Tappert, C.C.: A survey of binary similarity and
  distance measures.
\newblock Journal of Systemics, Cybernetics and Informatics \textbf{8}(1),
  43--48 (2010)

\bibitem{curve-area}
Dai, J., Chen, X., Chu, N.: Research on the extraction and classification of
  the concave point from fiber image.
\newblock In: IEEE 12th International Conference on Signal Processing (ICSP),
  pp. 709--712 (2014)

\bibitem{graphcut_twostage}
Daněk, O., Matula, P., Ortiz-de Solórzano, C., Muñoz-Barrutia, A., Maška,
  M., Kozubek, M.: Segmentation of touching cell nuclei using a two-stage graph
  cut model.
\newblock In: A.B. Salberg, J.~Hardeberg, R.~Jenssen (eds.) Image Analysis,
  \emph{Lecture Notes in Computer Science}, vol. 5575, pp. 410--419 (2009)

\bibitem{7952578}
Erdil, E., Mesadi, F., Tasdizen, T., Cetin, M.: Disjunctive normal shape
  boltzmann machine.
\newblock In: Proceedings of the IEEE International Conference on Acoustics,
  Speech and Signal Processing (ICASSP), pp. 2357--2361 (2017)

\bibitem{Eslami2014}
Eslami, S.M.A., Heess, N., Williams, C.K.I., Winn, J.: The shape boltzmann
  machine: A strong model of object shape.
\newblock International Journal of Computer Vision \textbf{107}(2), 155--176
  (2014)

\bibitem{farhan2013novel}
Farhan, M., Yli-Harja, O., Niemist{\"o}, A.: A novel method for splitting
  clumps of convex objects incorporating image intensity and using rectangular
  window-based concavity point-pair search.
\newblock Pattern Recognition \textbf{46}(3), 741--751 (2013)

\bibitem{fitzgibbon1999direct}
Fitzgibbon, A., Pilu, M., Fisher, R.B.: Direct least square fitting of
  ellipses.
\newblock IEEE Transactions on Pattern Analysis and Machine Intelligence
  \textbf{21}(5), 476--480 (1999)

\bibitem{hearst1998support}
Hearst, M.A., Dumais, S., Osman, E., Platt, J., Scholkopf, B.: Support vector
  machines.
\newblock Intelligent Systems and Their Applications \textbf{13}(4), 18--28
  (1998)

\bibitem{5518402}
Jung, C., Kim, C.: Segmenting clustered nuclei using h-minima transform-based
  marker extraction and contour parameterization.
\newblock IEEE Transactions onBiomedical Engineering \textbf{57}(10),
  2600--2604 (2010)

\bibitem{5193169}
Kothari, S., Chaudry, Q., Wang, M.: Automated cell counting and cluster
  segmentation using concavity detection and ellipse fitting techniques.
\newblock In: IEEE International Symposium on Biomedical Imaging, pp. 795--798
  (2009)

\bibitem{chord}
Kumar, S., Ong, S.H., Ranganath, S., Ong, T.C., Chew, F.T.: A rule-based
  approach for robust clump splitting.
\newblock Pattern Recognition \textbf{39}(6), 1088--1098 (2006)

\bibitem{5659480}
Law, Y.N., Lee, H.K., Liu, C., Yip, A.: A variational model for segmentation of
  overlapping objects with additive intensity value.
\newblock IEEE Transactions on Image Processing \textbf{20}(6), 1495--1503
  (2011)

\bibitem{6247778}
Lou, X., Koethe, U., Wittbrodt, J., Hamprecht, F.: Learning to segment dense
  cell nuclei with shape prior.
\newblock In: IEEE Conference on Computer Vision and Pattern Recognition
  (CVPR), pp. 1012--1018 (2012)

\bibitem{frs}
Loy, G., Zelinsky, A.: Fast radial symmetry for detecting points of interest.
\newblock IEEE Transactions on Pattern Analysis and Machine Intelligence
  \textbf{25}(8), 959--973 (2003)

\bibitem{1634509}
Mao, K.Z., Zhao, P., Tan, P.H.: Supervised learning-based cell image
  segmentation for p53 immunohistochemistry.
\newblock IEEE Transactions on Biomedical Engineering \textbf{53}(6),
  1153--1163 (2006)

\bibitem{martin2005use}
Martin, J.D., Simpson, T.W.: Use of kriging models to approximate deterministic
  computer models.
\newblock American Institute of Aeronautics and Astronautics Journal(AIAA J)
  \textbf{43}(4), 853--863 (2005)

\bibitem{meng1993ecm}
Meng, X.L., Rubin, D.B.: Maximum likelihood estimation via the ecm algorithm: A
  general framework.
\newblock Biometrika \textbf{80}(2), 267--278 (1993)

\bibitem{otsu1975threshold}
Otsu, N.: A threshold selection method from gray-level histograms.
\newblock Automatica \textbf{11}(285-296), 23--27 (1975)

\bibitem{npa}
Park, C., Huang, J.Z., Ji, J.X., Ding, Y.: Segmentation, inference and
  classification of partially overlapping nanoparticles.
\newblock IEEE Transactions on Pattern Analysis and Machine Intelligence
  \textbf{35}(3), 669--681 (2013)

\bibitem{prasad2012polygonal}
Prasad, D.K., Leung, M.K.: Polygonal representation of digital curves.
\newblock INTECH Open Access Publisher (2012)

\bibitem{sbfNuclei}
Quelhas, P., Marcuzzo, M., Mendonca, A., Campilho, A.: Cell nuclei and
  cytoplasm joint segmentation using the sliding band filter.
\newblock IEEE Transactions on Medical Imaging \textbf{29}(8), 1463--1473
  (2010)

\bibitem{RICHARDSON2017209}
Richardson, R.R., Osborne, M.A., Howey, D.A.: Gaussian process regression for
  forecasting battery state of health.
\newblock Journal of Power Sources \textbf{357}(Supplement C), 209 -- 219
  (2017)

\bibitem{rosenfeld1985}
Rosenfeld, A.: Measuring the sizes of concavities.
\newblock Pattern Recognition Letters \textbf{3}(1), 71--75 (1985)

\bibitem{bnd-skeleton}
Samma, A.S.B., Talib, A.Z., Salam, R.A.: Combining boundary and skeleton
  information for convex and concave points detection.
\newblock In: IEEE Seventh International Conference onComputer Graphics,
  Imaging and Visualization (CGIV), pp. 113--117 (2010)

\bibitem{ihc}
Shu, J., Fu, H., Qiu, G., Kaye, P., Ilyas, M.: Segmenting overlapping cell
  nuclei in digital histopathology images.
\newblock In: 35th International Conference on Medicine and Biology Society
  (EMBC), pp. 5445--5448 (2013)

\bibitem{7562400}
Song, Y., Tan, E.L., Jiang, X., Cheng, J.Z., Ni, D., Chen, S., Lei, B., Wang,
  T.: Accurate cervical cell segmentation from overlapping clumps in pap smear
  images.
\newblock IEEE Transactions on Medical Imaging \textbf{36}(1), 288--300 (2017)

\bibitem{cervical}
Song, Y., Zhang, L., Chen, S., Ni, D., Lei, B., Wang, T.: Accurate segmentation
  of cervical cytoplasm and nuclei based on multiscale convolutional network
  and graph partitioning.
\newblock IEEE Transactions on Biomedical Engineering \textbf{62}(10),
  2421--2433 (2015)

\bibitem{10.1007/978-3-319-24574-4_46}
Su, H., Xing, F., Kong, X., Xie, Y., Zhang, S., Yang, L.: Robust cell detection
  and segmentation in histopathological images using sparse reconstruction and
  stacked denoising autoencoders.
\newblock In: Proceedings of the International Conference on Medical Image
  Computing and Computer-Assisted Intervention (MICCAI), pp. 383--390. Springer
  International Publishing (2015)

\bibitem{skeleton}
Wang, W., Song, H.: Cell cluster image segmentation on form analysis.
\newblock In: IEEE Third International Conference on Natural Computation
  (ICNC), vol.~4, pp. 833--836 (2007)

\bibitem{curv-th}
Wen, Q., Chang, H., Parvin, B.: A delaunay triangulation approach for
  segmenting clumps of nuclei.
\newblock In: IEEE Sixth International Conference on Symposium on Biomedical
  Imaging: From Nano to Macro, ISBI'09, pp. 9--12. IEEE Press, Piscataway, NJ,
  USA (2009)

\bibitem{rasmussen2006gaussian}
Williams, C.K., Rasmussen, C.E.: Gaussian processes for machine learning.
\newblock MIT Press \textbf{2}(3), 4 (2006)

\bibitem{pmlr-v28-wilson13}
Wilson, A., Adams, R.: Gaussian process kernels for pattern discovery and
  extrapolation.
\newblock In: Proceedings of the 30th International Conference on Machine
  Learning, \emph{Proceedings of Machine Learning Research}, vol.~28, pp.
  1067--1075. PMLR (2013)

\bibitem{10.1007/978-3-319-24574-4_45}
Xie, Y., Kong, X., Xing, F., Liu, F., Su, H., Yang, L.: Deep voting: A robust
  approach toward nucleus localization in microscopy images.
\newblock In: Medical Image Computing and Computer-Assisted Intervention --
  MICCAI 2015. Springer International Publishing (2015)

\bibitem{7274740}
Xing, F., Xie, Y., Yang, L.: An automatic learning-based framework for robust
  nucleus segmentation.
\newblock IEEE Transactions on Medical Imaging \textbf{35}(2), 550--566 (2016)

\bibitem{6698355}
Xu, H., Lu, C., Mandal, M.: An efficient technique for nuclei segmentation
  based on ellipse descriptor analysis and improved seed detection algorithm.
\newblock IEEE Journal of Biomedical and Health Informatics \textbf{18}(5),
  1729--1741 (2014)

\bibitem{7274766}
{Yan}, J., {Li}, K., {Bai}, E., {Deng}, J., {Foley}, A.M.: Hybrid probabilistic
  wind power forecasting using temporally local gaussian process.
\newblock IEEE Transactions on Sustainable Energy \textbf{7}(1), 87--95 (2016)

\bibitem{YANG2018387}
Yang, D., Zhang, X., Pan, R., Wang, Y., Chen, Z.: A novel gaussian process
  regression model for state-of-health estimation of lithium-ion battery using
  charging curve.
\newblock Journal of Power Sources \textbf{384}, 387 -- 395 (2018)

\bibitem{YANG20142266}
Yang, H., Ahuja, N.: Automatic segmentation of granular objects in images:
  Combining local density clustering and gradient-barrier watershed.
\newblock Pattern Recognition \textbf{47}(6), 2266--2279 (2014)

\bibitem{YANG2018375}
Yang, J., Zhang, X., Wang, X.: Multi-scale shape boltzmann machine: A shape
  model based on deep learning method.
\newblock Procedia Computer Science \textbf{129}, 375--381 (2018)

\bibitem{concavity}
Yeo, T., Jin, X., Ong, S., Sinniah, R., et~al.: Clump splitting through
  concavity analysis.
\newblock Pattern Recognition Letters \textbf{15}(10), 1013--1018 (1994)

\bibitem{991040}
Yu, S.X., Shi, J.: Understanding popout through repulsion.
\newblock In: Proceedings of the IEEE Conference On the Computer Vision and
  Pattern Recognition (CVPR), vol.~2, pp. II--752--II--757 (2001)

\bibitem{Zafari2018}
Zafari, S.: Segmentation of partially overlapping convex objects in silhouette
  images.
\newblock Ph.D. thesis, Lappeenranta University of Technology (2018)

\bibitem{7300433}
Zafari, S., Eerola, T., Sampo, J., K{\"a}lvi{\"a}inen, H., Haario, H.:
  Segmentation of overlapping elliptical objects in silhouette images.
\newblock IEEE Transactions on Image Processing \textbf{24}(12), 5942--5952
  (2015)

\bibitem{zafari2015segmentation}
Zafari, S., Eerola, T., Sampo, J., K{\"a}lvi{\"a}inen, H., Haario, H.:
  Segmentation of partially overlapping nanoparticles using concave points.
\newblock In: Advances in Visual Computing, pp. 187--197. Springer (2015)

\bibitem{Zafari2017}
Zafari, S., Eerola, T., Sampo, J., K{\"a}lvi{\"a}inen, H., Haario, H.:
  Comparison of concave point detection methods for overlapping convex objects
  segmentation.
\newblock In: 20th Scandinavian Conference on Image Analysis:, SCIA 2017,
  Troms{\o}, Norway, June 12--14, 2017, Proceedings, Part II, pp. 245--256
  (2017)

\bibitem{Zafari2017BB}
Zafari, S., Eerola, T., Sampo, J., K{\"a}lvi{\"a}inen, H., Haario, H.:
  Segmentation of partially overlapping convex objects using branch and bound
  algorithm.
\newblock In: ACCV 2016 International Workshops, Taipei, Taiwan, November
  20-24, 2016, Revised Selected Papers, Part III, pp. 76--90 (2017)

\bibitem{cnn}
Zeiler, M.D., Fergus, R.: Visualizing and understanding convolutional networks.
\newblock In: D.~Fleet, T.~Pajdla, B.~Schiele, T.~Tuytelaars (eds.) Computer
  Vision -- ECCV 2014, pp. 818--833. Springer International Publishing, Cham
  (2014)

\bibitem{acos_1}
Zhang, Q., Pless, R.: Segmenting multiple familiar objects under mutual
  occlusion.
\newblock In: IEEE International Conference on Image Processing (ICIP), pp.
  197--200 (2006)

\bibitem{bubble}
Zhang, W.H., Jiang, X., Liu, Y.M.: A method for recognizing overlapping
  elliptical bubbles in bubble image.
\newblock Pattern Recognition Letters \textbf{33}(12), 1543--1548 (2012)

\end{thebibliography}

\end{document}